\documentclass[11pt,a4paper]{article}

\usepackage[a4paper,margin=2.2cm]{geometry}
\usepackage[T1]{fontenc}
\usepackage[utf8]{inputenc}
\usepackage{lmodern}
\usepackage{microtype}
\usepackage{amsmath,amssymb,amsthm,mathtools,bm}
\usepackage{booktabs,multirow,array,tabularx,longtable}
\usepackage{graphicx}
\usepackage{subcaption}
\usepackage{float}
\usepackage{xcolor}
\usepackage{hyperref}
\usepackage[nameinlink,noabbrev]{cleveref}
\usepackage{enumitem}
\usepackage{caption}
\usepackage{fancyhdr}
\usepackage{tikz}
\usepackage{pgfplots}
\usepackage[numbers,sort&compress]{natbib}

\definecolor{Accent}{RGB}{30,90,160}
\definecolor{Accent2}{RGB}{160,60,40}
\definecolor{GELU}{RGB}{255,140,0}
\definecolor{SiLU}{RGB}{128,0,128}
\definecolor{Dark}{RGB}{25,25,25}
\definecolor{Light}{RGB}{245,247,250}
\definecolor{Good}{RGB}{30,130,70}

\hypersetup{
    colorlinks=true,
    linkcolor=blue!60!black,
    citecolor=blue!60!black,
    urlcolor=blue!60!black
}

\newcolumntype{Y}{>{\raggedright\arraybackslash}X}
\newcommand{\dynact}{\text{dynAct}}
\newcommand{\baseact}{\text{BaseAct}}

\newcommand{\dynactmish}{\text{dynActivation(Mish)}}
\newcommand{\dynactgelu}{\text{dynActivation(GELU)}}
\newcommand{\dynactsilu}{\text{dynActivation(SiLU)}}
\newcommand{\dynactrelu}{\text{dynActivation(ReLU)}}

\title{\textbf{dynActivation: A Trainable Activation Family for Adaptive Nonlinearity }}
\author{Alois Bachmann\\Ruprecht-Karls-Universität Heidelberg\\\texttt{alois.bachmann@stud.uni-heidelberg.de}}
\date{21st of March 2026}

\pgfplotsset{compat=1.18} 

\begin{document}
\maketitle

\begin{abstract}
Standard activation functions impose a fixed nonlinearity on every layer of a neural network.
This paper proposes \emph{dynActivation}, a per-layer trainable activation defined as
$f_i(x) = \mathrm{BaseAct}(x)(\alpha_i - \beta_i) + \beta_i x$,
where $\alpha_i$ and $\beta_i$ are lightweight learned scalars that interpolate between the base nonlinearity and a linear path and $\mathrm{BaseAct}(x)$ resembles any ReLU-like function. The static and dynamic ReLU-like variants are then compared across multiple vision tasks, language modeling tasks, and ablation studies. The results suggest that dynActivation variants tend to linearize deep layers while maintaining high performance, which can improve training efficiency by up to $+54\%$ over ReLU..

On CIFAR-10~\cite{KrizhevskyCIFAR}, \dynactmish{} improves over static Mish~\cite{Misra2019Mish} by up to $+14.02\%$ on AttentionCNN~\cite{CNNA2021Attention} with an average improvment by $+6.00\%$, with a $24\%$ convergence-AUC reduction relative to Mish (2120 vs.\ 2785).
In a 1-to-75-layer MNIST~\cite{Deng2012MNIST} depth-scaling study, \dynact{} never drops below $95\%$ test accuracy ($95.3$--$99.3\%$), while ReLU collapses below $80\%$ at 25 layers.
Under FGSM~\cite{Goodfellow2015FGSM} at $\varepsilon{=}0.08$, \dynactmish{} incurs a $55.39\%$ accuracy drop versus $62.79\%$ for ReLU ($7.40\%$ advantage).
Transferred to language modeling, a new proposed dynActGLU(Swish)-variant achieves a $10.3\%$ relative perplexity reduction over SwiGLU~\cite{Shazeer2020GLUVariants} at 5{,}620 steps (4.047 vs.\ 4.514), though the gap vanishes at 34{,}300 steps.
\end{abstract}

\tableofcontents
\newpage

\section{Introduction}

Activation functions are a small but influential component of modern neural networks. Although they contribute few or no trainable parameters relative to the full model, they strongly affect gradient propagation, optimization stability, and final performance. In practice, however, activation functions are still chosen as fixed design decisions rather than learned components.

This paper proposes \emph{dynActivation}, a trainable activation family defined as $f_i(x) = \mathrm{BaseAct}(x)(\alpha_i - \beta_i) + \beta_i x$, where $\alpha_i$ and $\beta_i$ are per-layer scalars learned alongside the standard model weights.
This preserves the inductive bias of established base activations such as Mish, GELU, Swish, or ReLU while allowing each layer to adapt its effective nonlinearity.

One possible explanation for the improvements is that the additional linear path $\beta_i x$ appears to provide a gradient highway in layers where strong nonlinearity impedes information transport, while layers that benefit from high curvature retain it.

The experiments validate dynActivation through CIFAR~\cite{KrizhevskyCIFAR} image classification benchmarks, a 1-to-75-layer MNIST depth-scaling study, LLM transfer experiments, and a unified evaluation spanning adversarial robustness, optimizer stability, distribution-shift resilience, convergence, and ablations.

The main contributions are:
\begin{itemize}[leftmargin=1.4em]
    \item \textbf{Formulation and interpretation.} This work derives \emph{dynActivation} and provides a gradient-based interpretation showing how the linear path controlled by $\beta_i$ acts as a gradient highway.
    \item \textbf{CIFAR validation.} In a broad comparison \dynactmish{} achieves the highest mean accuracy on CIFAR-10 with statistical significant improvements over its static counterpart and up to improvements of $+14.02$ pp over static Mish~\cite{Misra2019Mish} on CIFAR-10/AttentionCNN and $+12.39$ pp over ReLU~\cite{nair2010relu} on CIFAR-100/AttentionCNN.
    \item \textbf{Depth scaling.} On MNIST, \dynact{} maintains $>95\%$ test accuracy from 1 to 75 layers, while ReLU collapses below $80\%$ at 25 layers.
    \item \textbf{Convergence.} \dynactmish{} reduces training-loss AUC by $24\%$ relative to static Mish (2120 vs.\ 2785).
    \item \textbf{Adversarial robustness.} Under FGSM~\cite{Goodfellow2015FGSM} at $\varepsilon{=}0.08$, \dynactmish{} reduces accuracy drop by $7.40$ pp versus ReLU, $8.28$ pp versus Mish, and $8.56$ pp versus Swish.
    \item \textbf{Optimizer--initialization stability.} Across 27 optimizer--init--learning-rate configurations on CIFAR-10/ResNet18~\cite{ResNet2016}, \dynactmish{} achieves mean accuracy $52.81\%$, outperforming Mish by $+3.34$ pp and ReLU by $+10.56$ pp.
    \item \textbf{LLM transfer.} dynActGLU(Swish) reduces perplexity by $10.3\%$ relative to SwiGLU~\cite{Shazeer2020GLUVariants} after 5{,}620 training steps, with no meaningful regression at 34{,}300 steps.
\end{itemize}

\section{Motivation}

\subsection{Why trainable activations}

A fixed activation imposes one global functional shape on all layers, regardless of their depth or role. Early, middle, and late layers may require different levels of nonlinearity, yet standard activations such as ReLU~\cite{nair2010relu}, GELU~\cite{Hendrycks2016GELU}, Swish~\cite{Ramachandran2017Swish}, and Mish~\cite{Misra2019Mish} remain fixed once selected. Recent work has further explored non-monotonic and smooth activation designs~\cite{Biswas2025WACV, SmoothActivations2021}. dynActivation lets the network adapt the effective shape of a chosen base activation during training while keeping the additional parameter count to two scalars per layer.

\subsection{Local intuition figure pair}

\begin{figure}[H]
\centering
\begin{subfigure}[t]{0.48\linewidth}
    \centering
    \resizebox{\linewidth}{!}{
    \begin{tikzpicture}[
      x=1cm,y=1cm,
      font=\sffamily\Large,
      >=stealth,
      neuron/.style={circle, draw=Dark, fill=white, minimum size=15pt, inner sep=0pt, line width=2.55pt},
      edge/.style={->, draw=gray!55, line width=2.40pt},
      actframe/.style={draw=black!25, rounded corners=2pt, line width=0.4pt, fill=white}
    ]
    
    \tikzset{
      pics/actaxes/.style={
        code={
          \draw[black!55, line width=2.05pt] (-1.0,0)--(1.0,0);
          \draw[black!55, line width=2.05pt] (0,-0.7)--(0,0.90);
        }
      },
      pics/mish/.style={
        code={
          \pic {actaxes};
          \draw[Accent, line width=3.20pt, domain=-3.35:3.35, samples=30, smooth] 
            plot (\x, { \x * tanh(ln(1+exp(\x))) });
        }
      }
    }
    
    \node[anchor=south, text=Accent, font=\bfseries\Large] at (4.2, 1.5)
      {Classic: Static non-linearity (e.g.: Mish)};
    
    \foreach \i/\yy in {1/0.8,2/0,3/-0.8}{
      \node[neuron] (I\i)  at (0, \yy) {};
      \node[neuron] (H1\i) at (2.1, \yy) {};
      \node[neuron] (H2\i) at (4.2, \yy) {};
      \node[neuron] (H3\i) at (6.3, \yy) {};
    }
    \node[neuron] (O) at (8.4, 0) {};
    
    \foreach \i in {1,2,3}{
      \foreach \j in {1,2,3}{
        \draw[edge] (I\i)--(H1\j);
        \draw[edge] (H1\i)--(H2\j);
        \draw[edge] (H2\i)--(H3\j);
      }
      \draw[edge] (H3\i)--(O);
    }
    
    \foreach \x/\p in {2.1/mish, 4.2/mish, 6.3/mish}{
      \node[actframe, minimum width=2.0cm, minimum height=1.75cm] at (\x,-2.05) {};
      \path (\x,-2.15) pic[scale=0.3] {\p};
    }
    
    \node[anchor=east] at (-0.5, 0) {Inp $\xrightarrow{}$};
    \node[anchor=west] at (8.9, 0) {$\xrightarrow{}$ Out};
    \end{tikzpicture}
    }
    \caption{Controlled deformation of a chosen base activation.}
\end{subfigure}
\hfill
\begin{subfigure}[t]{0.48\linewidth}
    \centering
    \resizebox{\linewidth}{!}{
    \begin{tikzpicture}[
      x=1cm,y=1cm,
      font=\sffamily\Large,
      >=stealth,
      neuron/.style={circle, draw=Dark, fill=white, minimum size=15pt, inner sep=0pt, line width=2.55pt},
      edge/.style={->, draw=gray!55, line width=2.40pt},
      actframe/.style={draw=black!25, rounded corners=2pt, line width=0.4pt, fill=white}
    ]
    
    \tikzset{
      pics/actaxes/.style={
        code={
          \draw[black!55, line width=2.05pt] (-1.0,0)--(1.0,0);
          \draw[black!55, line width=2.05pt] (0,-0.7)--(0,0.90);
        }
      },
      pics/dynA/.style={ 
        code={
          \pic {actaxes};
          \draw[Good, line width=3.20pt, domain=-3.35:3.35, samples=30, smooth] 
            plot (\x, { (\x * tanh(ln(1+exp(\x))))*-0.5 + 0.8*\x });
        }
      },
      pics/dynB/.style={ 
        code={
          \pic {actaxes};
          \draw[Good, line width=3.20pt, domain=-3.35:3.35, samples=30, smooth] 
            plot (\x, { (\x * tanh(ln(1+exp(\x))))*1.5 - 0.1*\x });
        }
      },
      pics/dynC/.style={ 
        code={
          \pic {actaxes};
          \draw[Good, line width=3.20pt, domain=-3.35:3.35, samples=30, smooth] 
            plot (\x, { (\x * tanh(ln(1+exp(\x))))*1.2 - 0.5*\x });
        }
      }
    }
    
    \node[anchor=south, text=Good, font=\bfseries\Large] at (4.2, 1.5)
      {dynActivation: Trained layer-specific non-linearity};
    
    \foreach \i/\yy in {1/0.8,2/0,3/-0.8}{
      \node[neuron] (I\i)  at (0, \yy) {};
      \node[neuron] (H1\i) at (2.1, \yy) {};
      \node[neuron] (H2\i) at (4.2, \yy) {};
      \node[neuron] (H3\i) at (6.3, \yy) {};
    }
    \node[neuron] (O) at (8.4, 0) {};
    
    \foreach \i in {1,2,3}{
      \foreach \j in {1,2,3}{
        \draw[edge] (I\i)--(H1\j);
        \draw[edge] (H1\i)--(H2\j);
        \draw[edge] (H2\i)--(H3\j);
      }
      \draw[edge] (H3\i)--(O);
    }
    
    \foreach \x/\p in {2.1/dynA, 4.2/dynB, 6.3/dynC}{
      \node[actframe, minimum width=2.0cm, minimum height=1.75cm] at (\x,-2.05) {};
      \path (\x,-2.15) pic[scale=0.3] {\p};
    }
    
    \node[anchor=east] at (-0.5, 0) {Inp $\xrightarrow{}$};
    \node[anchor=west] at (8.9, 0) {$\xrightarrow{}$ Out};
    \end{tikzpicture}
    }
    \caption{Different layers can learn different effective nonlinearities.}
\end{subfigure}
\caption{Motivational view of trainable activations.}
\label{fig:motivation_pair}
\end{figure}
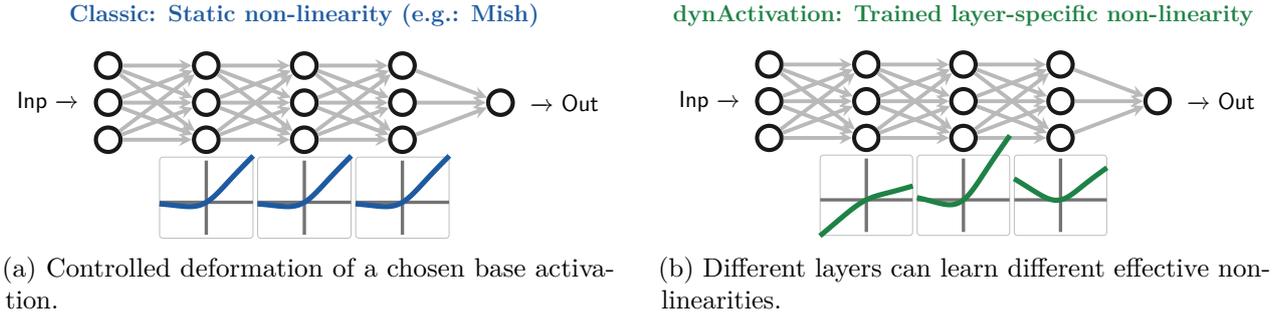

\section{dynActivation}

\subsection{Derivation}

A piecewise-linear activation can be interpreted as having two arms, one for negative and one for positive inputs. Allowing these arms to move independently would let the activation adapt its effective slope, but a hard piecewise construction would be non-smooth and less suitable for gradient-based optimization.

To avoid this, the construction begins with a smooth sigmoid-based gating mechanism. After combining both branches and simplifying, the result takes the form of a Swish-like term plus an explicitly linear term. Replacing the sigmoid-based base term by a general $\mathrm{BaseAct}(x)$ turns the construction into a family of trainable activations.

The resulting activation for layer \(i\) is
\begin{equation}
f_i(x) := \mathrm{BaseAct}(x)\cdot(\alpha_i-\beta_i)+\beta_i x.
\end{equation}
This formulation keeps the base activation as the main nonlinear component while allowing two trainable parameters to adapt its effective contribution and its linear correction.

\paragraph{Derivation:}
\begin{align}
\sigma(x) &= \frac{1}{1+e^{-x}} \\
\mathrm{dynSigmoid}_{\mathrm{right}}(x) &= \sigma(x)\alpha \\
\mathrm{dynSigmoid}_{\mathrm{left}}(x) &= (\sigma(x)-1)\beta \\
\mathrm{dynActivation}(x) &= \sigma(x)x(\alpha-\beta)+\beta x \\
&\Rightarrow \baseact(x)(\alpha-\beta)+\beta x
\end{align}

The special case $\mathrm{dynAct}(x):=\mathrm{dynActivation}(x;\mathrm{BaseAct}=\mathrm{Mish})$ serves as the default variant in several experiments, motivated by its strong average performance and stability in the comparative benchmarks.

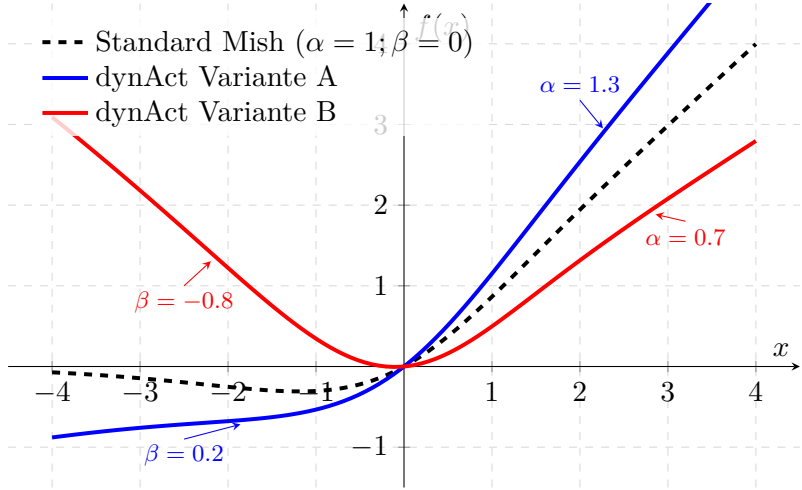
\begin{figure}[H]
\centering

\begin{tikzpicture}
    \begin{axis}[
        axis lines=middle,
        xlabel={$x$},
        ylabel={$f(x)$},
        domain=-4:4,
        samples=100,
        ymin=-1.5, ymax=4.5,
        xmin=-4.5, xmax=4.5,
        legend pos=north west,
        legend style={draw=none, fill=white, fill opacity=0.8, text opacity=1},
        legend cell align={left},
        grid=major,
        grid style={dashed, gray!30},
        width=12cm,
        height=8cm,
        declare function={
            mish(\x) = \x * tanh(ln(1+exp(\x)));
            dynAct(\x,\a,\b) = mish(\x) * (\a - \b ) + (\x) * (\b);
        }
    ]
    
    \addplot [black, dashed, line width=1.5pt] {mish(x)};
    \addlegendentry{Standard Mish ($\alpha=1 ; \beta = 0$)}
    
    \addplot [blue, line width=1.5pt] {dynAct(x, 1.3, 0.2)};
    \addlegendentry{dynAct Variante A}
    
    \addplot [red, line width=1.5pt] {dynAct(x, 0.7, -0.8)};
    \addlegendentry{dynAct Variante B}
    
    \node[blue, font=\footnotesize] at (axis cs:-2.5, -1.1) {$\beta = 0.2$};
    \draw[blue, ->, >=stealth, shorten >=2pt] (axis cs:-2.5, -0.9) -- (axis cs:-1.8, -0.7);
    
    \node[red, font=\footnotesize] at (axis cs:-2.5, 0.8) {$\beta = -0.8$};
    \draw[red, ->, >=stealth, shorten >=2pt] (axis cs:-2.5, 1.0) -- (axis cs:-2.15, 1.35);
    
    \node[blue, font=\footnotesize] at (axis cs:2.0, 3.5) {$\alpha = 1.3$};
    \draw[blue, ->, >=stealth, shorten >=2pt] (axis cs:2.0, 3.3) -- (axis cs:2.3, 2.9);
    
    \node[red, font=\footnotesize] at (axis cs:3.2, 1.6) {$\alpha = 0.7$};
    \draw[red, ->, >=stealth, shorten >=2pt] (axis cs:3.2, 1.8) -- (axis cs:2.8, 1.9);
    
    \end{axis}
\end{tikzpicture}

\caption{Schematic view of the dynActivation family and its shape deformation relative to the base activation.}
\label{fig:method_family}
\end{figure}

\begin{table}[H]
\centering
\caption{Core definitions of the dynActivation family.}
\label{tab:method_definitions}
\begin{tabularx}{\linewidth}{>{\raggedright\arraybackslash}p{3.2cm}Y}
\toprule
Object & Definition / interpretation \\
\midrule
    $\mathrm{BaseAct}(x)$ & Chosen base activation, e.g. Mish, GELU, Swish, or ReLU. \\
$\alpha_i$ & Trainable per-layer coefficient scaling the base activation contribution. \\
$\beta_i$ & Trainable per-layer coefficient defining the explicit linear path. \\
$f_i(x)$ & $\mathrm{BaseAct}(x)\cdot(\alpha_i-\beta_i)+\beta_i x$. \\
$\mathrm{dynAct}(x)$ & Special case using Mish as base activation. \\
\bottomrule
\end{tabularx}
\end{table}

\subsection{Parameter interpretation}

The parameter \(\alpha_i\) controls how strongly the base activation contributes to the resulting function, whereas \(\beta_i\) introduces an explicit linear component. When \(\alpha_i \approx 1\) and \(\beta_i \approx 0\), dynActivation stays close to the original base activation. When \(\alpha_i \approx \beta_i\), the nonlinear contribution is reduced and the function becomes approximately linear.

\subsection{Gradient and parameter analysis}

The gradient structure of dynActivation provides the key intuition for why the activation may improve training behavior. From the definition, the derivative yields
\begin{align}
\frac{\partial f_i}{\partial x} &= \mathrm{BaseAct}'(x)\cdot(\alpha_i-\beta_i)+\beta_i, \\
\frac{\partial f_i}{\partial \alpha_i} &= \mathrm{BaseAct}(x), \\
\frac{\partial f_i}{\partial \beta_i} &= x-\mathrm{BaseAct}(x).
\end{align}

The additive \(\beta_i\) term in \(\partial f_i/\partial x\) creates a linear gradient path independent of the base activation's derivative. The model can therefore learn a more direct transport path whenever the optimization problem benefits from it, offering a plausible explanation for improved convergence or greater stability in deeper networks.

The update for \(\alpha_i\) is driven directly by the base activation output, while the update for \(\beta_i\) depends on the difference between the identity path and the base activation. Initializing \(\beta_i=0\) does not freeze the parameter, because its gradient does not vanish at that initialization.

\subsection{Gradient-flow / Lipschitz figure pair}

\begin{figure}[H]
\centering
\begin{subfigure}[t]{0.58\linewidth}
    \centering
    \includegraphics[width=\linewidth]{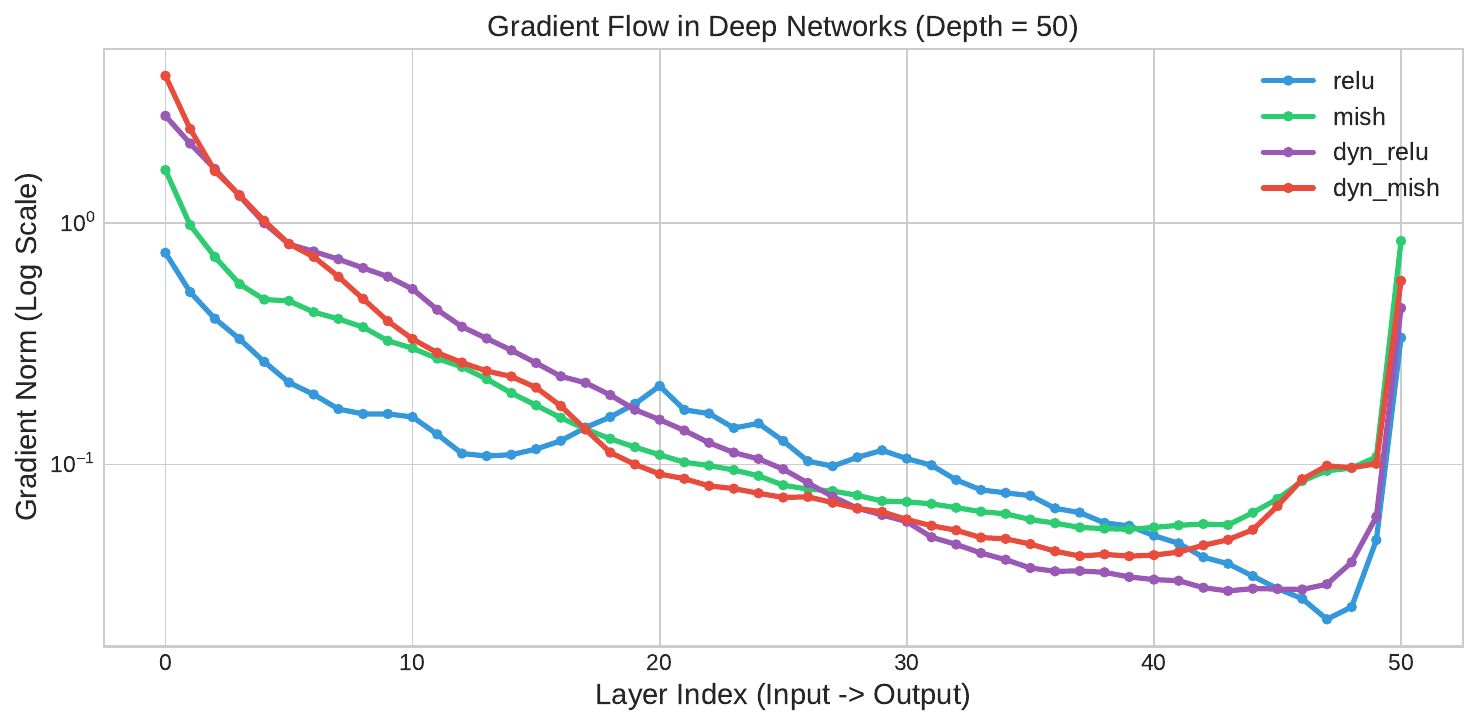}
    \caption{Gradient flow over normalized depth.}
\end{subfigure}
\hfill
\begin{subfigure}[t]{0.38\linewidth}
    \centering
    \includegraphics[width=\linewidth]{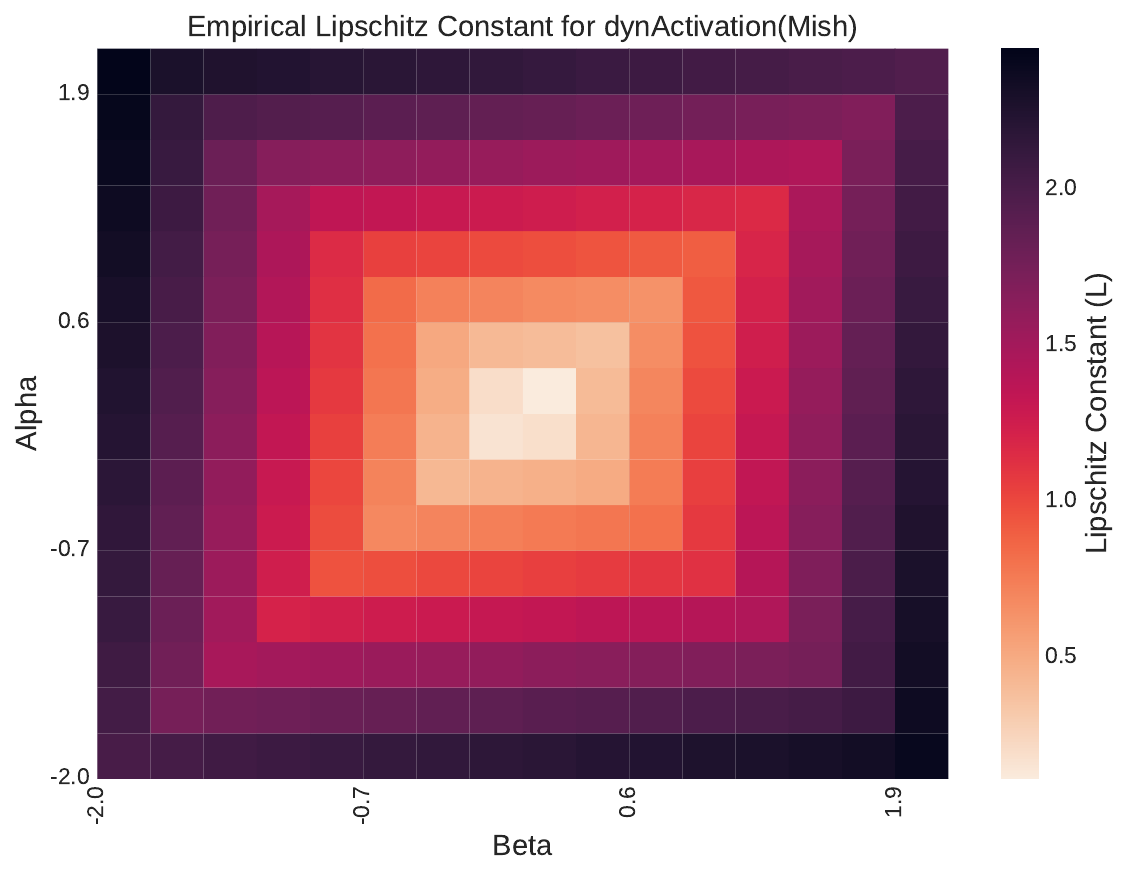}
    \caption{Lipschitz behavior over parameter settings.}
\end{subfigure}
\caption{Gradient flow and Lipschitz behavior of dynActivation.}
\label{fig:theory_pair}
\end{figure}

Figure~\ref{fig:theory_pair} (left) provides a qualitative comparison of gradient-flow behavior across depth in a 50-layer network. Compared with the static activations, the dynamic variants show less severe gradient decay through much of the network, especially beyond the early layers. This suggests that the learned linear component can help preserve gradient propagation, although the effect is best interpreted qualitatively rather than through exact ratios read from the plot.

Figure~\ref{fig:theory_pair} (right) shows the empirical Lipschitz landscape of dynActivation as a function of $(\alpha,\beta)$. The plot indicates a broad region of moderate Lipschitz values surrounded by larger values toward the edges of the parameter space. The parameters reached during training fall within this moderate region, supporting the claim that the learned activation typically operates away from obviously unstable settings.

\section{Experimental Setup}

\subsection{Protocol}

All experiments follow a reproducible protocol with fixed seeds, explicitly reported training settings, and architecture-specific benchmark configurations. Classification results are reported using test accuracy and test loss, while additional modules report metrics such as  corruption robustness, adversarial robustness, convergence AUC, and efficiency measurements.

 \begin{table}[H]
\centering

\begin{tabular}{ccccc}
\toprule
System & GPU (VRAM) & CPU & RAM & SSD \\
\midrule
    System 1 & RTX 2080 Ti (11 GB) & Ryzen 7 5700X CPU & 32 GB DDR4 & 1 TB \\
    System 2 (Cloud) & RTX 5090 (32 GB) & AMD EPYC 7763 & 112.8 GB DDR4 & 16 GB \\

\bottomrule

\end{tabular}
\caption{Hardware specification of Training Systems.}
\label{tab:hardware}
\end{table}

\subsection{Architectures and datasets}

The evaluation uses CIFAR-10 and CIFAR-100~\cite{KrizhevskyCIFAR} as the main image-classification benchmarks. MNIST~\cite{Deng2012MNIST} serves for the depth-scaling and overfitting analysis because its relative simplicity makes depth-dependent degradation easier to observe. Language-model transfer is tested in a Monad-70M~\cite{PleIAsMonad} setting on the SYNTH dataset through a dynActGLU formulation. The unified suite combines results from robustness, calibration, efficiency, and ablation modules across multiple architectures.

The compared activations include both standard and dynamic variants. In the main CIFAR comparisons, the paper considers ReLU, GELU, SiLU, Mish~\cite{Misra2019Mish}, and their dynamic counterparts. The extended comparison broadens this set to include ELU~\cite{Clevert2016ELU}, PReLU~\cite{He2015PReLU}, SELU~\cite{Klambauer2017SELU}, and others.

\section{CIFAR Validation}

\subsection{Environment Setup}

\begin{table}[H]
\centering

\begin{tabularx}{\linewidth}{>{\raggedright\arraybackslash}p{4cm}Y}
\toprule
Setting & Value \\
\midrule
    Framework & PyTorch \\
    Optimizer & Adam~\cite{Kingma2015Adam} \\
    Learning rate & 0.001 \\
    Weight decay & 5e-9 \\
    Scheduler & StepLR(step\_size=10, gamma=0.1) \\
    Gradient clipping & clip\_grad\_norm(..., max\_norm=1.0) \\
    CIFAR batch size & 128 \\
    MNIST batch size & 256 \\
    CIFAR epochs & 30 \\
    Statistical runs & 5 seeds per architecture-dataset-activation combination \\
    Augmentation & RandomCrop(32, padding=4), RandomHorizontalFlip, dataset-specific normalization \\    
\bottomrule

\end{tabularx}
\caption{Experimental protocol and reproducibility settings.}
\label{tab:protocol}
\end{table}

\subsection{dynActivation variants comparison}

Table~\ref{tab:dyn_variant_avg} reports the average performance across all seven dataset--architecture combinations.
\dynactmish{} leads with a mean accuracy of $66.86\%$ ($\sigma = 0.58$), followed by \dynactgelu{} at $66.69\%$ ($-0.17$ pp deficit), \dynactsilu{} at $65.65\%$, and \dynactrelu{} at $60.91\%$ ($-5.95$ pp).
The lowest standard deviation for \dynactmish{} ($\sigma = 0.58$ vs.\ $1.99$ for \dynactrelu{}) indicates that the Mish base not only achieves the highest mean but is also the most consistent across configurations.
These results identify \dynactmish{} as the recommended default within the dynActivation family, while leaving open the possibility of local improvements from other base activations on specific architectures.

\begin{table}[H]
\centering

\begin{tabular}{lcccc}
\toprule
Activation & Mean Acc & Std Acc & Mean Loss & Std Loss \\
\midrule
    \dynactmish{} & 66.8571 & 0.5845 & 1.1781 & 0.0235 \\
\dynactgelu{} & 66.6886 & 0.7935 & 1.1810 & 0.0312 \\
\dynactsilu{} & 65.6500 & 1.1292 & 1.2304 & 0.0365 \\
\dynactrelu{} & 60.9117 & 1.9969 & 1.4215 & 0.0998 \\
\bottomrule
\end{tabular}
\caption{Average performance over all dataset-architecture combinations for the dynamic variants.}
\label{tab:dyn_variant_avg}
\end{table}

\subsection{Best-per-combination test}

Table~\ref{tab:best_dyn_per_combo} reports the best dynamic activation separately for each dataset--architecture combination.

\dynactmish{} wins 5 out of 7 dataset--architecture combinations.
The performance gains range from $+0.79$ pp on CIFAR-10/MobileNet~\cite{Howard2017MobileNet} (over GELU) to $+14.02$ pp on CIFAR-10/AttentionCNN over static Mish and $+4.63$ pp over ReLU on the same combination.
On CIFAR-10/DenseNet~\cite{Huang2017DenseNet}, \dynactmish{} improves ReLU by $+1.76$ pp.
\dynactgelu{} is locally superior on CIFAR-10/ResNet18 and CIFAR-100/AttentionCNN, the latter achieving $45.70\%$ vs.\ $33.31\%$ for ReLU ($+12.39$ pp), the single largest margin in the study.

\begin{table}[H]
\centering

\begin{tabular}{lllcc}
\toprule
Dataset & Architecture & Best dynActivation & Mean Acc & Mean Loss \\
\midrule
    CIFAR10 & AttentionCNN & \dynactmish{} & 78.27 & 0.6390 \\
CIFAR10 & DenseNet & \dynactmish{} & 81.92 & 0.5466 \\
CIFAR10 & MobileNet & \dynactmish{} & 80.87 & 0.5743 \\
CIFAR10 & ResNet18 & \dynactgelu{} & 84.11 & 0.5196 \\
CIFAR100 & AttentionCNN & \dynactgelu{} & 45.70 & 2.0433 \\
CIFAR100 & DenseNet & \dynactmish{} & 52.42 & 1.7713 \\
CIFAR100 & MobileNet & \dynactmish{} & 45.39 & 2.1282 \\
\bottomrule
\end{tabular}
\caption{Best dynamic variant per dataset-architecture combination.}
\label{tab:best_dyn_per_combo}
\end{table}

\subsection{Full comparison}

The comparison emphasizes both mean accuracy and seed-to-seed dispersion.
The single largest gain appears on CIFAR-100/AttentionCNN, where \dynactgelu{} reaches $45.70\%$ versus $33.31\%$ for ReLU, a margin of $+12.39$ pp.
On CIFAR-10/AttentionCNN, \dynactmish{} at $78.27\%$ exceeds static Mish ($64.25\%$, $-14.02$ pp) and ReLU ($73.64\%$, $-4.63$ pp) while also reducing standard deviation from $1.43$ to $0.85$.
Across the full table, every dynamic variant improves upon at least one static baseline in its winning combination, and none exhibits the variance inflation seen in \dynactrelu{} at the variant level.

\begin{longtable}{l c c c c c c}
\caption{CIFAR-Statistics: Comparisons of the activations per combination.
\(\Delta\)Acc/\(\Delta\)Loss are relative to the best dynActivation-variant for the specific combination.}
\label{tab:cifar_full_per_combo}\\
\toprule
Aktivierung & Mean Acc (\%) & $\sigma$(Acc) & Mean Loss & $\sigma$(Loss) & $\Delta$Acc & $\Delta$Loss \\
\midrule
\endfirsthead

\toprule
Aktivierung & Mean Acc (\%) & $\sigma$(Acc) & Mean Loss & $\sigma$(Loss) & $\Delta$Acc & $\Delta$Loss \\
\midrule
\endhead

\midrule
\multicolumn{7}{r}{Continuation on the next page.}\\
\midrule
\endfoot

\bottomrule
\endlastfoot

\midrule
\multicolumn{7}{l}{\textbf{CIFAR10 -- AttentionCNN}} \\
\midrule
ReLU & 73.64 & 1.4321 & 0.7605 & 0.0497 & -4.63 & +0.1215 \\
GELU & 69.83 & 1.7864 & 0.8624 & 0.0542 & -8.44 & +0.2234 \\
Mish & 64.25 & 1.6702 & 1.0001 & 0.0508 & -14.02 & +0.3611 \\
SiLU & 63.95 & 1.2492 & 1.0086 & 0.0323 & -14.32 & +0.3696 \\
\textbf{\dynactmish{}} & \textbf{78.27} & \textbf{0.8523} & \textbf{0.6390} & \textbf{0.0259} & +0.00 & +0.0000 \\

\midrule
\midrule
\multicolumn{7}{l}{\textbf{CIFAR10 -- DenseNet}} \\
\midrule
ReLU & 80.16 & 0.6478 & 0.5868 & 0.0212 & -1.76 & +0.0402 \\
GELU & 80.83 & 0.9647 & 0.5637 & 0.0330 & -1.09 & +0.0171 \\
Mish & 80.20 & 0.7894 & 0.5888 & 0.0258 & -1.72 & +0.0422 \\
SiLU & 80.45 & \textbf{0.4493} & 0.5835 & 0.0205 & -1.47 & +0.0369 \\
\textbf{\dynactmish{}} & \textbf{81.92} & \textbf{0.4494} & \textbf{0.5466} & \textbf{0.0151} & +0.00 & +0.0000 \\

\midrule
\midrule
\multicolumn{7}{l}{\textbf{CIFAR10 -- MobileNet}} \\
\midrule
ReLU & 77.93 & 1.4169 & 0.6719 & 0.0582 & -2.94 & +0.0976 \\
GELU & 80.08 & 0.5638 & 0.5886 & \textbf{0.0134} & -0.79 & +0.0143 \\
Mish & 79.16 & 0.6710 & 0.6209 & 0.0246 & -1.71 & +0.0466 \\
SiLU & 79.05 & \textbf{0.3782} & 0.6212 & 0.0138 & -1.82 & +0.0469 \\
\textbf{\dynactmish{}} & \textbf{80.87} & 0.4648 & \textbf{0.5743} & 0.0178 & +0.00 & +0.0000 \\

\midrule
\midrule
\multicolumn{7}{l}{\textbf{CIFAR10 -- ResNet18}} \\
\midrule
ReLU & 83.54 & 0.2272 & 0.4991 & \textbf{0.0059} & -0.57 & -0.0205 \\
GELU & 84.07 & \textbf{0.1581} & 0.4988 & 0.0093 & -0.04 & -0.0208 \\
Mish & 83.97 & 0.2868 & \textbf{0.4920} & 0.0086 & -0.14 & -0.0276 \\
SiLU & 83.95 & 0.1927 & 0.4935 & 0.0063 & -0.16 & -0.0261 \\
\dynactmish{} & 84.02 & 0.3222 & 0.5276 & 0.0106 & -0.09 & +0.0080 \\
\textbf{\dynactgelu{}} & \textbf{84.11} & 0.3765 & 0.5196 & 0.0111 & +0.00 & +0.0000 \\

\midrule
\midrule
\multicolumn{7}{l}{\textbf{CIFAR100 -- AttentionCNN}} \\
\midrule
ReLU & 33.31 & 2.5522 & 2.5215 & 0.1082 & -12.39 & +0.4782 \\
GELU & 30.69 & 1.3821 & 2.6836 & 0.0852 & -15.01 & +0.6403 \\
Mish & 27.55 & 1.0424 & 2.8212 & 0.0649 & -18.15 & +0.7779 \\
SiLU & 24.65 & \textbf{0.7931} & 2.9693 & 0.0345 & -21.05 & +0.9260 \\
\dynactmish{} & 45.55 & 0.9498 & 2.0463 & \textbf{0.0472} & -0.15 & +0.0030 \\
\textbf{\dynactgelu{}} & \textbf{45.70} & 1.0976 & \textbf{2.0433} & 0.0552 & +0.00 & +0.0000 \\

\midrule
\midrule
\multicolumn{7}{l}{\textbf{CIFAR100 -- DenseNet}} \\
\midrule
ReLU & 51.18 & 1.0539 & 1.8240 & 0.0399 & -1.24 & +0.0527 \\
GELU & 51.16 & 0.9901 & 1.8170 & 0.0341 & -1.26 & +0.0457 \\
Mish & 48.06 & 1.9885 & 1.9529 & 0.1048 & -4.36 & +0.1816 \\
SiLU & 48.61 & 0.8971 & 1.9248 & 0.0426 & -3.81 & +0.1535 \\
\dynactmish{} & 51.98 & \textbf{0.4402} & 1.7846 & \textbf{0.0124} & -0.44 & +0.0133 \\
\textbf{\dynactgelu{}} & \textbf{52.42} & 0.7582 & \textbf{1.7713} & 0.0319 & +0.00 & +0.0000 \\

\midrule
\midrule
\multicolumn{7}{l}{\textbf{CIFAR100 -- MobileNet}} \\
\midrule

ReLU & 40.99 & 1.6794 & 2.2493 & 0.0794 & -4.40 & +0.1211 \\
GELU & 44.07 & 0.9372 & \textbf{2.1049} & 0.0451 & -1.32 & -0.0233 \\
Mish & 42.82 & 0.7421 & 2.1509 & 0.0316 & -2.57 & +0.0227 \\
SiLU & 42.24 & \textbf{0.2910} & 2.1806 & \textbf{0.0134} & -3.15 & +0.0524 \\
\textbf{\dynactmish{}} & \textbf{45.39} & 0.6126 & 2.1282 & 0.0353 & +0.00 & +0.0000 \\

\end{longtable}

\subsection{Extended activation comparison}

To avoid drawing conclusions from too narrow a baseline set, the comparison extends to a comprehensive group of 27 standard and recent trainable activations—including variants such as Apa, MELU, PReLU, and TAAF. 

As reported in Table \ref{tab:activation_results}, dynActivation(Mish) achieves the highest overall accuracy at $79.72 \pm 0.0073$, improving on its static base activation Mish ($78.91 \pm 0.0041$, a +0.81 pp improvement) and outperforming recent trainable alternatives like Apa ($79.26\%$) and Trainable\_swish ($79.11\%$). The dynActivation family claims the top three ranks overall, with dynActivation(GELU) and dynActivation(SiLU) achieving $79.69$ and $79.52$ respectively. Notably, the lowest-performing dynamic variant, dynActivation(ReLU) at $78.22$, still outperforms static ReLU ($77.66$).

\begin{table}[H]
\centering
\begin{tabular}{l c c c}
\toprule
\textbf{Activation} & \textbf{Accuracy (Mean $\pm$ Std)} & \textbf{Loss (Mean $\pm$ Std)} & \textbf{Trainable} \\
\midrule
Apa~\cite{Pikoulis2024APA} & $0.7926 \pm 0.0055$ & $0.6287 \pm 0.0178$ & Yes \\
Trainable\_swish~\cite{Ramachandran2017Swish} & $0.7911 \pm 0.0047$ & $0.6318 \pm 0.0112$ & Yes \\
Pdelu~\cite{Cheng2020PDELU} & $0.7909 \pm 0.0054$ & $0.6474 \pm 0.0217$ & Yes \\
Melu~\cite{Nanni2022MeLU} & $0.7904 \pm 0.0075$ & $0.6597 \pm 0.0258$ & Yes \\
Mish~\cite{Misra2019Mish} & $0.7891 \pm 0.0041$ & $0.6348 \pm 0.0108$ & No \\
Prelu\_scalar~\cite{He2015PReLU} & $0.7874 \pm 0.0057$ & $0.6740 \pm 0.0136$ & No \\
Taaf~\cite{TAAF2024} & $0.7874 \pm 0.0047$ & $0.6654 \pm 0.0180$ & Yes \\
Adaptive\_gelu~\cite{Hendrycks2016GELU} & $0.7871 \pm 0.0044$ & $0.6449 \pm 0.0143$ & Yes \\
SiLU~\cite{Elfwing2018SiLU} & $0.7869 \pm 0.0037$ & $0.6368 \pm 0.0109$ & No \\
Apl~\cite{Agostinelli2014APL} & $0.7861 \pm 0.0099$ & $0.6774 \pm 0.0274$ & Yes \\
GELU~\cite{Hendrycks2016GELU} & $0.7836 \pm 0.0053$ & $0.6545 \pm 0.0169$ & No \\
Hardswish~\cite{Howard2019MobileNetV3} & $0.7828 \pm 0.0040$ & $0.6460 \pm 0.0101$ & No \\
ELU~\cite{Clevert2016ELU} & $0.7814 \pm 0.0032$ & $0.6368 \pm 0.0107$ & No \\
CELU~\cite{Barron2017CELU} & $0.7814 \pm 0.0032$ & $0.6368 \pm 0.0107$ & No \\
Laaf\_relu~\cite{Jagtap2020LAAF} & $0.7785 \pm 0.0066$ & $0.6608 \pm 0.0183$ & Yes \\
ReLU~\cite{nair2010relu} & $0.7766 \pm 0.0033$ & $0.6615 \pm 0.0102$ & No \\
LeakyReLU~\cite{Maas2013LeakyReLU} & $0.7765 \pm 0.0055$ & $0.6582 \pm 0.0150$ & No \\
Srelu~\cite{Jin2015SReLU} & $0.7677 \pm 0.0079$ & $0.7042 \pm 0.0244$ & Yes \\
SELU~\cite{Klambauer2017SELU} & $0.7660 \pm 0.0049$ & $0.6776 \pm 0.0146$ & No \\
Softplus~\cite{Dugas2001Softplus} & $0.7630 \pm 0.0074$ & $0.6798 \pm 0.0186$ & No \\
Saaf~\cite{Yang2021SAAF} & $0.7556 \pm 0.0062$ & $0.7297 \pm 0.0190$ & Yes \\
Tanh~\cite{LeCun1998Tanh} & $0.7330 \pm 0.0031$ & $0.7723 \pm 0.0094$ & No \\
Erfrelu~\cite{Rajanand2023ErfReLU} & $0.7170 \pm 0.0073$ & $0.8428 \pm 0.0203$ & Yes \\
\midrule
dynActivation(Mish) & $0.7972 \pm 0.0073$ & $0.6720 \pm 0.0214$ & Yes \\
dynActivation(GELU) & $0.7969 \pm 0.0060$ & $0.6709 \pm 0.0189$ & Yes \\
dynActivation(SiLU) & $0.7952 \pm 0.0040$ & $0.6738 \pm 0.0154$ & Yes \\
dynActivation(ReLU) & $0.7822 \pm 0.0031$ & $0.6944 \pm 0.0158$ & Yes \\
\bottomrule
\end{tabular}
\caption{Comparison of standard and dynAct variants, sorted descending by mean accuracy.}
\label{tab:activation_results}
\end{table}

\subsection{Statistical significance tests}

Table \ref{tab:significance_results} details the statistical significance of the performance differences measured against the leading activation, dynActivation(Mish). Using a paired t-test over the aggregated combination results, dynActivation(Mish) demonstrates a statistically significant advantage ($p < 0.05$) over the vast majority of the 26 competing functions, including static base activations like Mish ($p = 0.0375$), ReLU ($p = 0.0205$), and closely related trainable variants such as MELU ($p = 0.0017$) and TAAF ($p = 0.0466$).

The statistical test fails to reject the null hypothesis only for a small cluster of highly competitive activations. Specifically, dynActivation(GELU) ($p = 0.9280$) and dynActivation(SiLU) ($p = 0.4485$) are statistically indistinguishable from the Mish variant, confirming that the dynActivation framework provides robust benefits regardless of the exact base function. Among external activations, Apa ($p = 0.2598$), Trainable\_swish ($p = 0.0581$), and Adaptive\_gelu ($p = 0.0758$) perform competitively enough that their difference from the top rank is not strictly significant at the $0.05$ threshold, though dynActivation(Mish) maintains a higher absolute mean accuracy in all cases.

\begin{table}[H]
\centering
\begin{tabular}{l c c c}
\toprule
\textbf{Activation} & \textbf{$p$-Value} & \textbf{$t$-Statistic} & \textbf{Significant vs. Best} \\
\midrule
Apa~\cite{Pikoulis2024APA} & 0.2598 & -1.2026 & No \\
Trainable\_swish~\cite{Ramachandran2017Swish} & 0.0581 & -2.1703 & No \\
Pdelu~\cite{Cheng2020PDELU} & 0.1188 & -1.7238 & No \\
Melu~\cite{Nanni2022MeLU} & 0.0017 & -4.4033 & Yes \\
Mish~\cite{Misra2019Mish} & 0.0375 & -2.4373 & Yes \\
Prelu\_scalar~\cite{He2015PReLU} & 0.0208 & -2.7976 & Yes \\
Taaf~\cite{TAAF2024} & 0.0466 & -2.3055 & Yes \\
Adaptive\_gelu~\cite{Hendrycks2016GELU} & 0.0758 & -2.0060 & No \\
SiLU~\cite{Elfwing2018SiLU} & 0.0649 & -2.1020 & No \\
Apl~\cite{Agostinelli2014APL} & 0.0474 & -2.2952 & Yes \\
GELU~\cite{Hendrycks2016GELU} & 0.0535 & -2.2207 & No \\
Hardswish~\cite{Howard2019MobileNetV3} & 0.0406 & -2.3893 & Yes \\
ELU~\cite{Clevert2016ELU} & 0.0019 & -4.3338 & Yes \\
CELU~\cite{Barron2017CELU} & 0.0019 & -4.3345 & Yes \\
Laaf\_relu~\cite{Jagtap2020LAAF} & 0.0386 & -2.4202 & Yes \\
ReLU~\cite{nair2010relu} & 0.0205 & -2.8050 & Yes \\
LeakyReLU~\cite{Maas2013LeakyReLU} & 0.0156 & -2.9732 & Yes \\
Srelu~\cite{Jin2015SReLU} & 0.0002 & -5.9098 & Yes \\
SELU~\cite{Klambauer2017SELU} & $4.20 \times 10^{-5}$ & -7.3831 & Yes \\
Softplus~\cite{Dugas2001Softplus} & $5.70 \times 10^{-5}$ & -7.0963 & Yes \\
Saaf~\cite{Yang2021SAAF} & $1.10 \times 10^{-5}$ & -8.7252 & Yes \\
Tanh~\cite{LeCun1998Tanh} & $0.00 \times 10^{0}$ & -18.3604 & Yes \\
Erfrelu~\cite{Rajanand2023ErfReLU} & 0.0013 & -4.6186 & Yes \\
\midrule
dynActivation(Mish) & - & - & No \\
dynActivation(GELU) & 0.9280 & -0.0929 & No \\
dynActivation(SiLU) & 0.4485 & -0.7925 & No \\
dynActivation(ReLU) & 0.0140 & -3.0425 & Yes \\
\bottomrule
\end{tabular}
\caption{Statistical significance testing results against \dynactmish{}, sorted ascending by p-value.}
\label{tab:significance_results}
\end{table}

\subsection{Local figure + local interpretation}

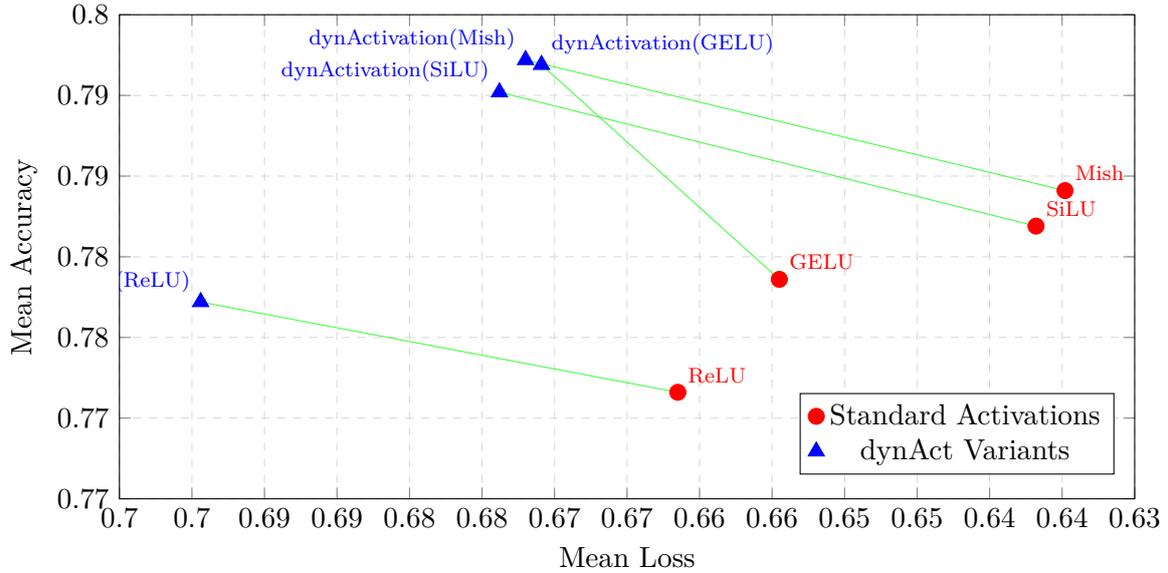
\begin{figure}[H]
\centering
\begin{tikzpicture}
\begin{axis}[
width=0.9\linewidth,
height=8cm,
xlabel={Mean Loss},
ylabel={Mean Accuracy},
x dir=reverse,
grid=major,
grid style={dashed, gray!30},
legend style={at={(0.67,0.05)}, anchor=south west},
xmin=0.63, xmax=0.7,
ymin=0.77, ymax=0.8,
]

\addlegendentry{Standard Activations}

\addplot[
only marks,
mark=*,
color=red,
mark size=3pt
] coordinates {
(0.6348, 0.7891) 
(0.6368, 0.7869) 
(0.6545, 0.7836) 
(0.6615, 0.7766) 
};
\addlegendentry{dynAct Variants}

\addplot[
only marks,
mark=triangle*,
color=blue,
mark size=3.5pt
] coordinates {
(0.6720, 0.7972) 
(0.6738, 0.7952) 
(0.6709, 0.7969) 
(0.6944, 0.7822) 
};

\node[anchor=south west, font=\scriptsize, text=red] at (axis cs:0.6348, 0.7891) {Mish};
\node[anchor=south west, font=\scriptsize, text=red] at (axis cs:0.6368, 0.7869) {SiLU};
\node[anchor=south west, font=\scriptsize, text=red] at (axis cs:0.6545, 0.7836) {GELU};
\node[anchor=south west, font=\scriptsize, text=red] at (axis cs:0.6615, 0.7766) {ReLU};

\node[anchor=south east, font=\scriptsize, text=blue] at (axis cs:0.6720, 0.7972) {\dynactmish{}};
\node[anchor=south east, font=\scriptsize, text=blue] at (axis cs:0.6738, 0.7952) {\dynactsilu{}};
\node[anchor=south west, font=\scriptsize, text=blue] at (axis cs:0.6709, 0.7969) {\dynactgelu{}};
\node[anchor=south east, font=\scriptsize, text=blue] at (axis cs:0.6944, 0.7822) {\dynactrelu{}};

\draw[->, green!75] (axis cs:0.6348, 0.7891) -- (axis cs:0.6720, 0.7972);
\draw[->, green!75] (axis cs:0.6368, 0.7869) -- (axis cs:0.6738, 0.7952);
\draw[->, green!75] (axis cs:0.6545, 0.7836) -- (axis cs:0.6709, 0.7969);
\draw[->, green!75] (axis cs:0.6615, 0.7766) -- (axis cs:0.6944, 0.7822);

\end{axis}
\end{tikzpicture}
\caption{Accuracy-versus-Loss trade-off.}
\label{fig:acc_loss_tradeoff}
\end{figure}

Figure~\ref{fig:acc_loss_tradeoff} plots each activation in accuracy-versus-loss space and connects each base activation to its dynamic counterpart. 
The plot reveals that dynamizing a base activation increases its final mean accuracy across the evaluated configurations. \dynactmish{} achieves the highest accuracy at $79.72\%$, displacing static Mish ($78.91\%$) as the most accurate variant overall. 

However, unlike the accuracy gains, the dynamic variants exhibit a slightly higher test loss compared to their static baselines (e.g., $0.6720$ for \dynactmish{} versus $0.6348$ for Mish). This indicates a trade-off where the dynamic variants achieve a higher absolute correct classification rate, but may produce slightly more confident incorrect predictions or slightly less confident correct predictions.

\section{Overfitting Analysis on MNIST}

\subsection{Experimental setup for MNIST depth scaling}

The MNIST analysis studies how model performance changes as network depth increases, using progressively deeper networks with two convolutional front-end layers and fully connected layers of fixed width. Batch normalization~\cite{IoffeSzegedy2015BatchNorm} and dropout~\cite{Srivastava2014Dropout} are included to reduce trivial training failures.

\subsection{Depth scaling figure pair}

\begin{figure}[H]
\centering
\begin{subfigure}[t]{0.48\linewidth}
    \centering
    \includegraphics[width=\linewidth]{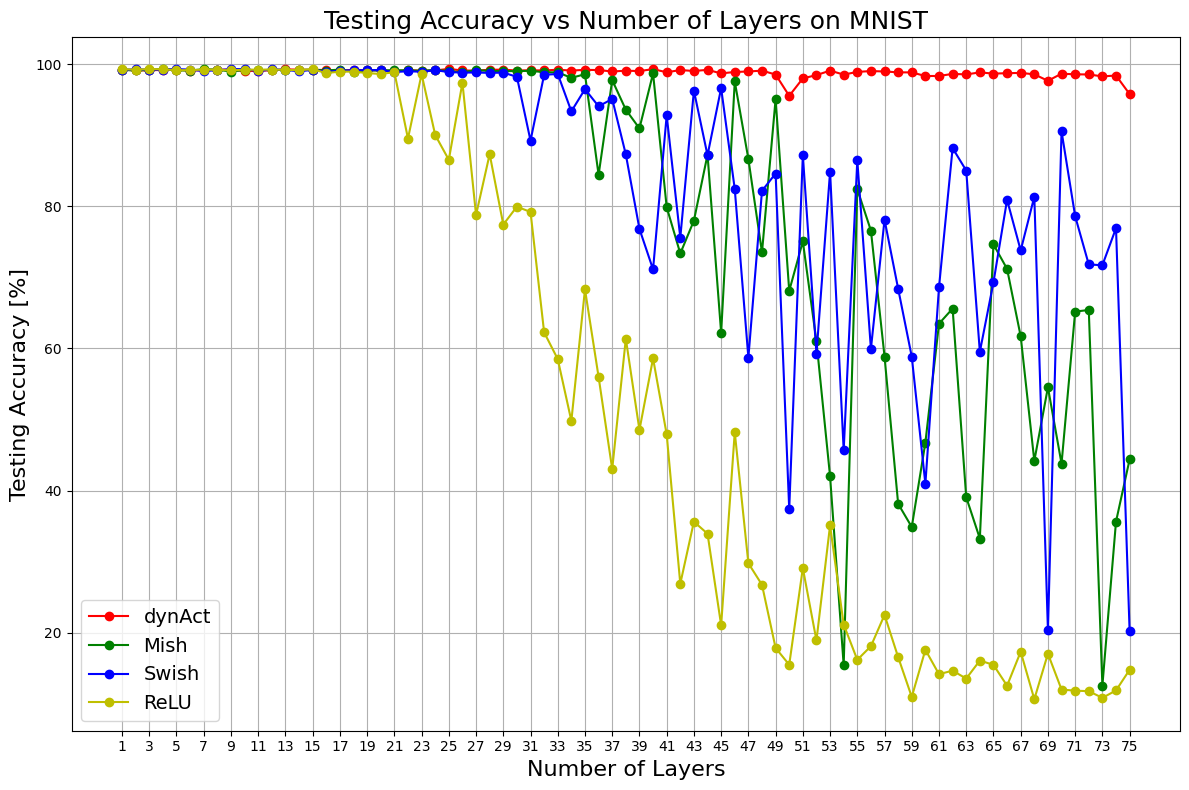}
    \caption{Test accuracy vs. depth.}
\end{subfigure}
\hfill
\begin{subfigure}[t]{0.48\linewidth}
    \centering
    \includegraphics[width=\linewidth]{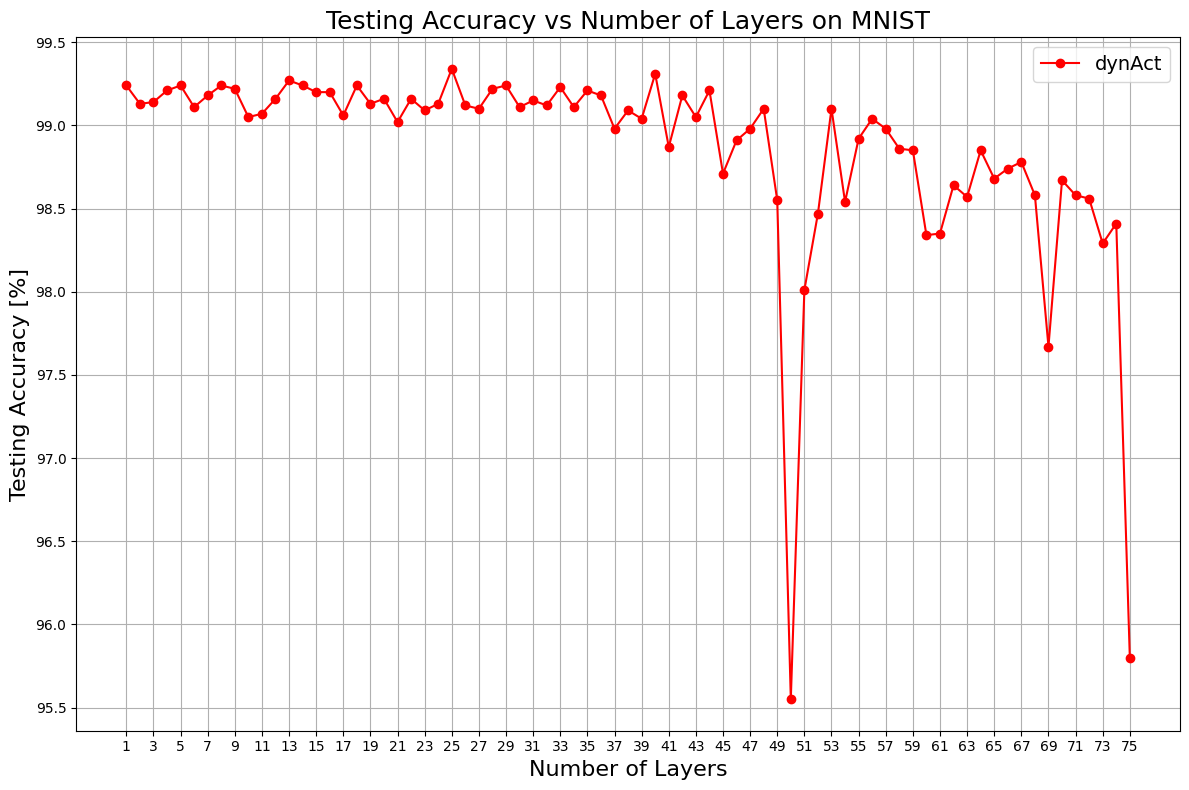}
    \caption{Test accuracy vs. depth. (only \dynact{})}
\end{subfigure}
\caption{MNIST depth scaling figure pair.}
\label{fig:mnist_depth_pair}
\end{figure}

Figure~\ref{fig:mnist_depth_pair} (left) plots test accuracy from 1 to 75 layers for ReLU, Swish, Mish, and \dynact{}.
\dynact{} (Mish base) is the only activation that never drops below $95\%$ across the entire sweep, ranging between $95.3\%$ and $99.3\%$.
In contrast, ReLU collapses below $80\%$ at approximately 25 layers and falls as low as $10$--$20\%$ beyond 45 layers.
Mish maintains near-$99\%$ accuracy up to about 15 layers before oscillating; it drops to approximately $60\%$ at layers 40--50 and fluctuates between $40$--$80\%$ at 50--75 layers.
Swish begins degrading around 25--30 layers, oscillating between $40$--$80\%$ and reaching approximately $40\%$ by layer 75.
The zoomed panel (right) sees some degradation of \dynact{} which seem negligible in comparison to the accuracy drop of the other activations.

\subsection{Learned activation visualization}

\begin{figure}[H]
\vspace{-1em}
\centering
\includegraphics[width=0.86\linewidth]{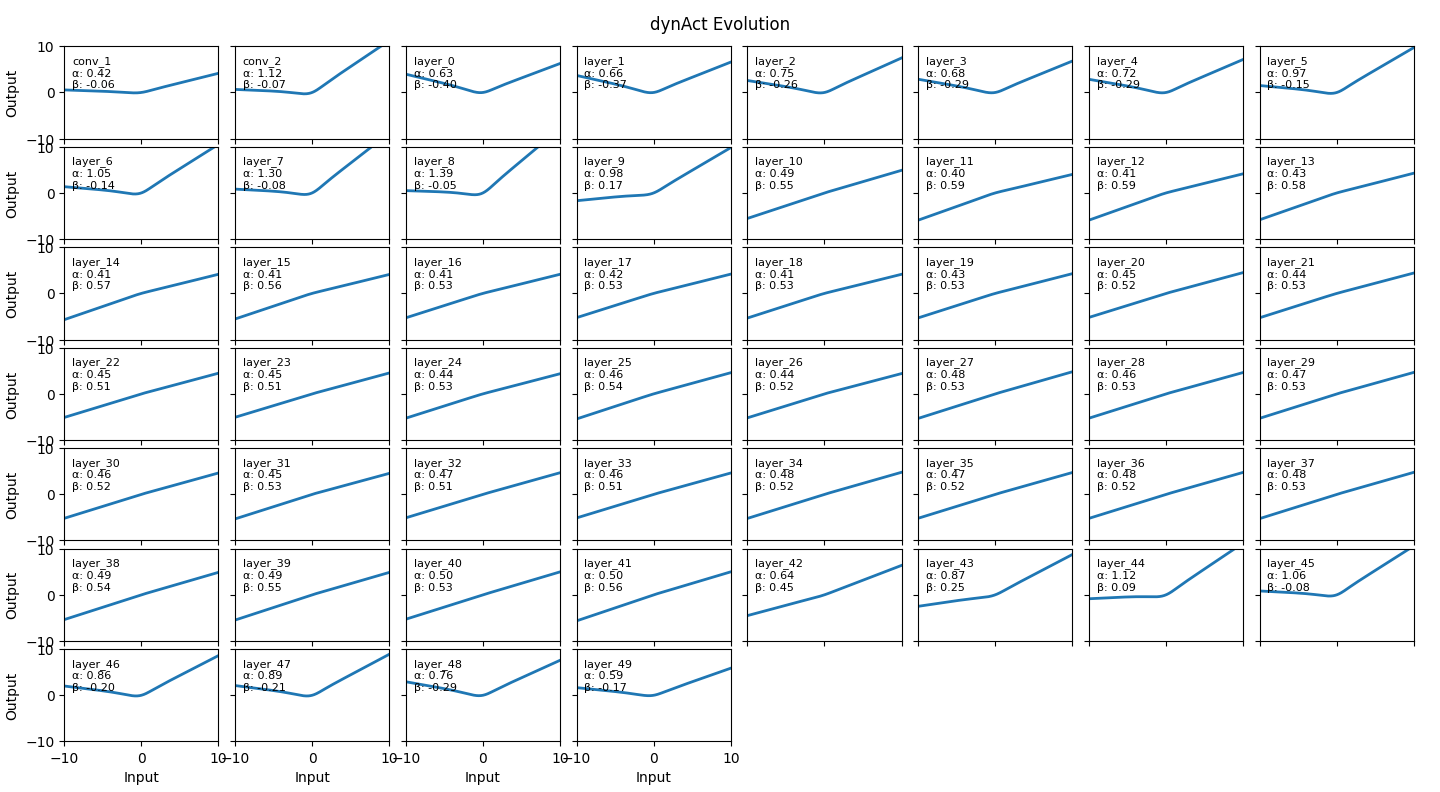}
\caption{Learned layer-wise activation shapes for a 50-layer MNIST network.}
\label{fig:mnist_shapes}
\vspace{-1em}
\end{figure}

Figure~\ref{fig:mnist_shapes} displays the learned per-layer activation shapes for the 50-layer MNIST network.
Early convolutional layers adopt shapes close to the Mish base activation: \texttt{conv\_1} converges to $\alpha = 0.42$, $\beta = -0.06$; \texttt{conv\_2} to $\alpha = 1.12$, $\beta = -0.07$.
The early fully connected layers (layers 6--8) sustain strong nonlinearity with $\alpha$ values of 1.05--1.39 and $\beta$ near $-0.06$ to $-0.14$.
In the middle layers (14--30) a clear trend emerges: $\alpha$ decreases to 0.41--0.60 while $\beta$ increases to 0.51--0.57, pushing the activation toward near-linear behavior.
Deep layers (30--45) consolidate at $\alpha \approx 0.45$--$0.50$, $\beta \approx 0.51$--$0.55$, retaining only a residual curvature from the Mish base.
The final layers (46--49) display diverse specialized shapes with $\alpha$ from 0.06 to 0.76 and $\beta$ from $-0.20$ to $0.39$, suggesting task-specific fine-tuning near the output.
This systematic transition from high-$\alpha$/low-$\beta$ (Mish-like) in early layers to high-$\beta$/low-$\alpha$ (near-linear) in deep layers is a quantitative expression of selective nonlinearity allocation.
This way the network behaves more like a model of the size of 20 layers which in turn then seems reasonable why it performs so well, because all the activations perform favourably at this size.

\subsection{Interpretation}

The MNIST results provide direct evidence for the selective nonlinearity hypothesis.
In the 50-layer network, the network transitions from Mish-like activations ($\alpha \approx 1.1$, $\beta \approx -0.07$) in early layers to near-linear behavior ($\alpha \approx \beta \approx 0.5$) in middle and deep layers.
Layers where $\alpha \approx \beta$ satisfy $f_i(x) \approx \beta_i x$, which simplifies to a near-identity map for moderate input values, effectively creating learned skip connections within the activation.
This mechanism reduces the effective curvature of deep layers without requiring explicit residual connections, which may explain the observed absence of depth-induced accuracy collapse.
Notably, \dynact{} is the only activation maintaining $>95\%$ accuracy at 75 layers, consistent with its ability to adapt its depth-wise nonlinearity profile.

\section{LLM Transfer Experiments}

\subsection{dynActGLU formulation}

To test whether the dynActivation idea transfers beyond image models, this work adapts it to the feedforward block of a language model and proposes a new dynActGLU formulation following similar intuition as before. The static intermediate nonlinearity in the baseline GLU-type layer is replaced by a dynamic variant built on the same base function, preserving architectural comparability while isolating the effect of trainable activation behavior.

\begin{align}
    \mathrm{FFN}_{\mathrm{dynActGLU}}(x, W_1, V_\alpha, V_\beta, W_2)
    &= \bigl((\mathrm{BaseAct}(xW_1)\odot(xV_\alpha - xV_\beta) + xW_1 \odot xV_\beta) \odot xW_1\bigr)\,W_2.
\end{align}

\subsection{Short-run table}

Table~\ref{tab:llm_short} reports validation loss and perplexity after 5{,}620 training steps.

\begin{table}[H]
\centering
\caption{Short-run LLM results after 5620 training steps.}
\label{tab:llm_short}
\begin{tabular}{lcc}
\toprule
Activation & Loss & Perplexity \\
\midrule
    Baseline SwiGLU~\cite{Shazeer2020GLUVariants} & 1.5072 & 4.5142 \\
dynActGLU(Swish) & 1.3979 & 4.0467 \\
\bottomrule
\end{tabular}
\end{table}

\subsection{Long-run table}

Table~\ref{tab:llm_long} reports validation loss and perplexity after 34{,}300 training steps.

\begin{table}[H]
\centering
\caption{Long-run LLM results after 34300 training steps.}
\label{tab:llm_long}
\begin{tabular}{lcc}
\toprule
Activation & Loss & Perplexity \\
\midrule
    Baseline SwiGLU & 1.3374 & 3.8091 \\
dynActGLU(Swish) & 1.3364 & 3.8056 \\
\bottomrule
\end{tabular}
\end{table}

\section{Unified Evaluation}

\subsection{Stability tests}

\subsubsection{Initialization stability}

The initialization-stability analysis tests different starting values for \(\alpha\) and \(\beta\) and reports both mean performance and instability.

\begin{figure}[H]
\centering
\includegraphics[width=0.86\linewidth]{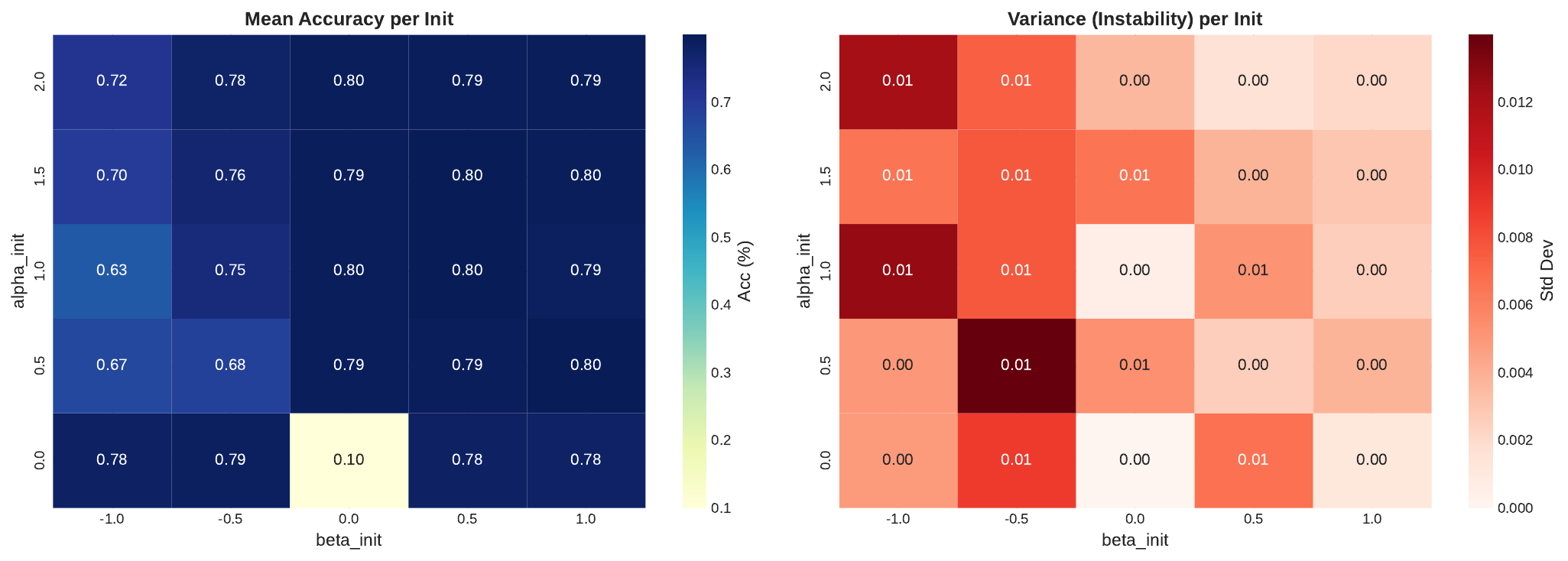}
\caption{Initialization-ablation heatmap over $\alpha_{{init}}$ and $\beta_{{init}}$.}
\label{fig:unified_init_heatmap}
\end{figure}

Figure~\ref{fig:unified_init_heatmap} confirms that the default initialization $\alpha_{\text{init}}=1$, $\beta_{\text{init}}=0$ achieves peak accuracy of $0.80$. The same peak is reached at $(\alpha=2, \beta=0)$, and the majority of configurations yield accuracy in the range $0.78$--$0.80$, demonstrating broad robustness to initialization. The only catastrophic failure occurs at $(\alpha=0, \beta=0)$, where accuracy collapses to $0.10$---consistent with the activation degenerating to the zero function when both parameters vanish. Variance remains below $0.01$ for most configurations, with a slightly elevated cluster near $(\alpha=0.5, \beta=-0.5)$.

\subsubsection{Optimizer-/init-stability test}

Across 27 configurations (3 initialization schemes $\times$ 3 optimizers $\times$ 3 learning rates) on CIFAR-10/ResNet18~\cite{ResNet2016}, \dynactmish{} achieves the highest mean accuracy at $52.81\%$ ($\sigma = 25.18$), followed by Mish~\cite{Misra2019Mish} at $49.47\%$ ($+3.34$ pp advantage) and ReLU~\cite{nair2010relu} at $42.25\%$ ($+10.56$ pp advantage). The best single configuration for \dynactmish{} reaches $75.77\%$ (kaiming\_normal initialization, RMSProp optimizer~, $\text{lr}=0.001$).

\begin{figure}[H]
\centering
\includegraphics[width=\linewidth]{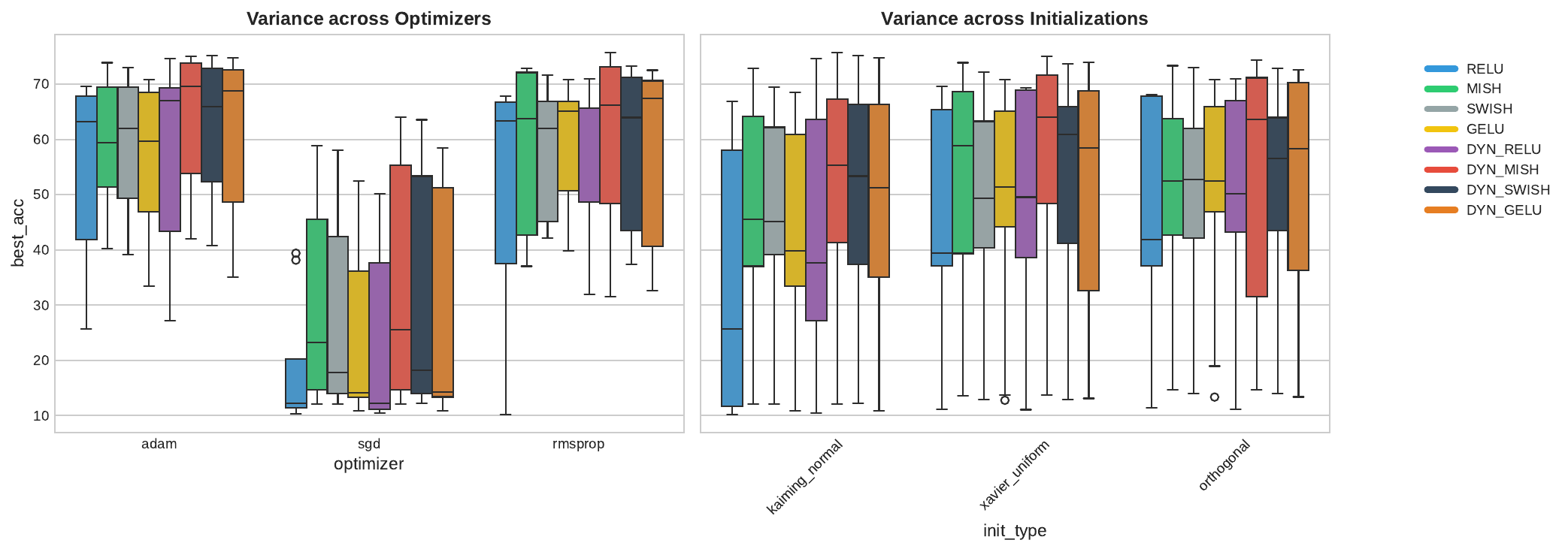}

\caption{Variance across optimizers and initialization schemes.}
\label{fig:unified_optimizer_var}
\end{figure}

\subsubsection{Hyperparameter-stability test}

dynActivation is tested on multiple learning rates, batch sizes and weight decay levels to test for hyperparameter robustness so that the a good run is not just a lucky run. 

It can be shown that the dynActivation-family can maintain its performance advantage despite different hyperparameters suggesting a good robustness against variations in them.

\begin{figure}[H]
    \centering
    \includegraphics[width=\linewidth]{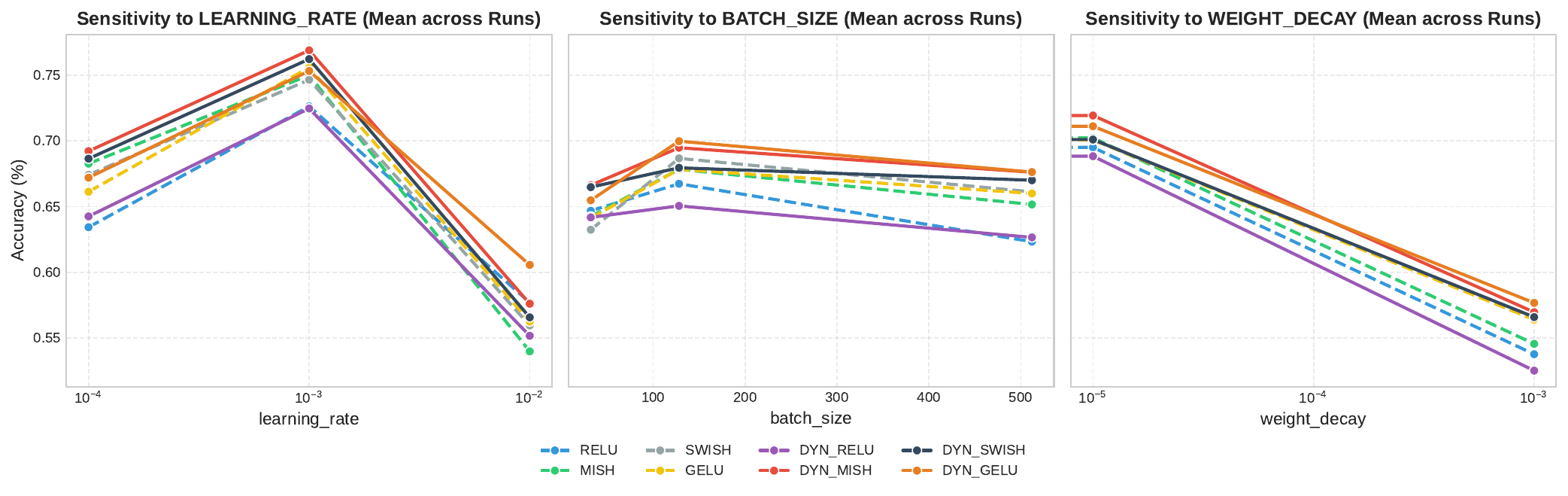}

    \caption{Robustness across hyperparameters.}
    \label{fig:hyperparam_iso}
\end{figure}

\subsubsection{Monotonicity / curvature test}

\begin{figure}[H]
\centering
\includegraphics[width=0.86\linewidth]{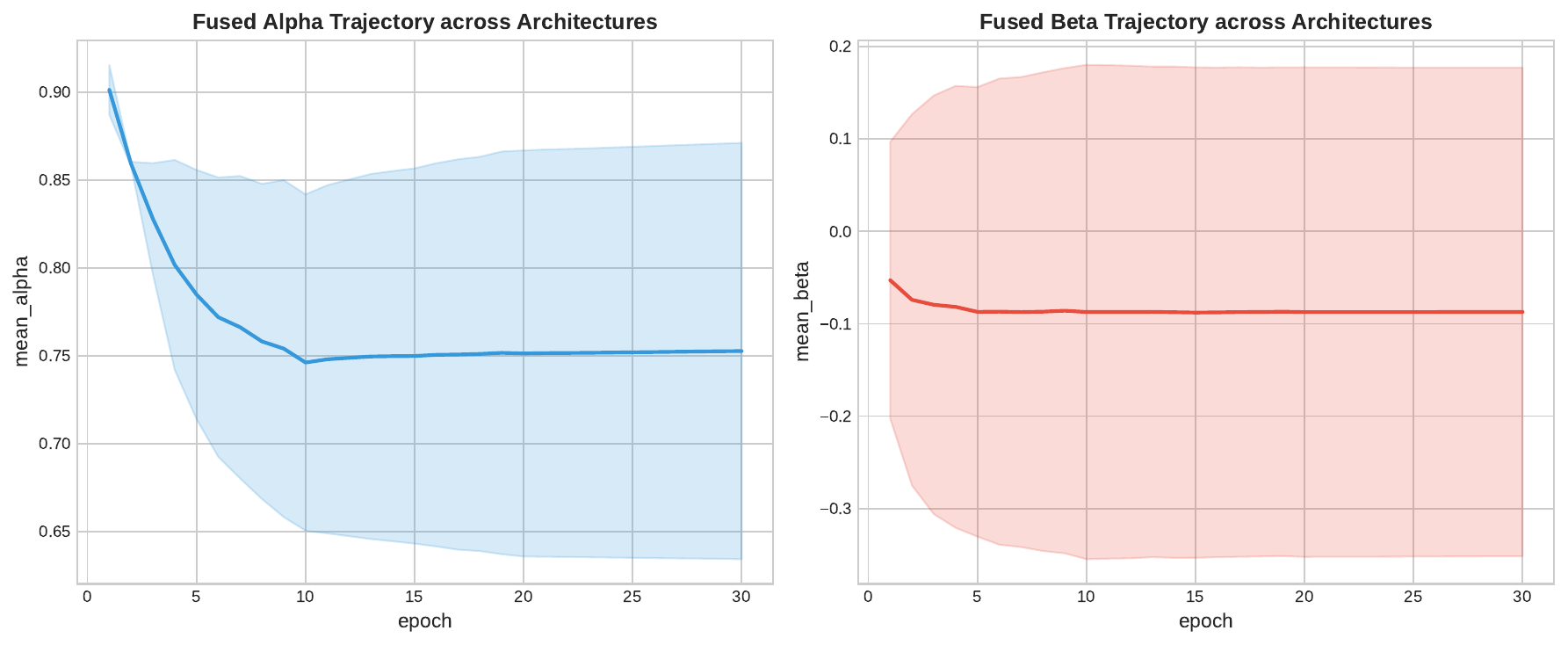}
\caption{Trajectories of mean and variance of activation parameters across epochs.}
\label{fig:unified_monotonicity}
\end{figure}

Figure~\ref{fig:unified_monotonicity} shows the trajectories of mean and variance of $\alpha$ and $\beta$ over 30 training epochs.
$\bar{\alpha}$ starts near $0.90$ at epoch 1, drops sharply to $\approx 0.80$ by epoch 5, and converges to $\approx 0.75$ by epoch 10--15, where it remains stable.
The variance band for $\alpha$ is initially narrow, widens to a range of approximately $0.65$--$0.87$ during epochs 5--10, then stabilizes at $\pm 0.10$ around the mean.
$\bar{\beta}$ starts near $-0.05$ at epoch 1, shifts to $\approx -0.08$ by epoch 5, and stabilizes there through epoch 30.
The variance band for $\beta$ is substantially larger ($\pm 0.2$ around the mean throughout training), indicating that different layers specialize their linear-path parameter to a much greater degree than the nonlinearity-scaling parameter.
Both parameters reach their converged regime by approximately epoch 10--15, demonstrating rapid and stable adaptation with no ongoing drift.

\subsection{Robustness tests}

\subsubsection{Distribution-shift test}

\begin{figure}[H]
\centering
\includegraphics[width=\linewidth]{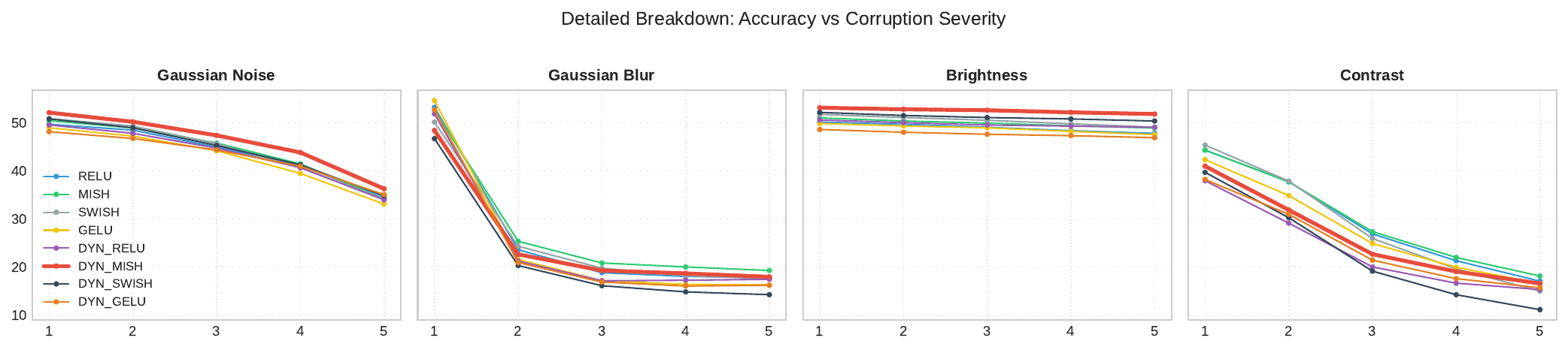}
\caption{Distribution-shift robustness across corruption types and severities.}
\label{fig:unified_distribution_shift}
\end{figure}

Figure~\ref{fig:unified_distribution_shift} plots accuracy across four corruption types~\cite{Hendrycks2019Robustness} (Gaussian noise, Gaussian blur, brightness, contrast) at five severity levels.
Under Gaussian noise at severity 1, \dynactmish{} leads with $52.14\%$, outperforming ReLU ($49.72\%$) by $+2.42$ pp and Mish ($50.26\%$) by $+1.88$ pp.
Under brightness perturbation at severity 5, \dynactmish{} achieves $51.84\%$, the highest among all activations, exceeding ReLU by $+4.02$ pp ($47.82\%$).
However, under contrast corruption, \dynactmish{} falls behind: at severity 1, Swish leads at $45.38\%$ while \dynactmish{} reaches only $41.00\%$ ($-4.38$ pp); at severity 5, Mish achieves $18.14\%$ versus $16.58\%$ for \dynactmish{}.
The overall pattern is mixed: \dynactmish{} shows clear advantages under noise and brightness corruptions, while exhibiting a moderate disadvantage under contrast perturbations. The additional trainable parameters do not introduce a uniform vulnerability but shift the degradation profile in a corruption-type-dependent manner.

\subsubsection{Adversarial-robustness test}

\begin{figure}[H]
\centering
\includegraphics[width=0.86\linewidth]{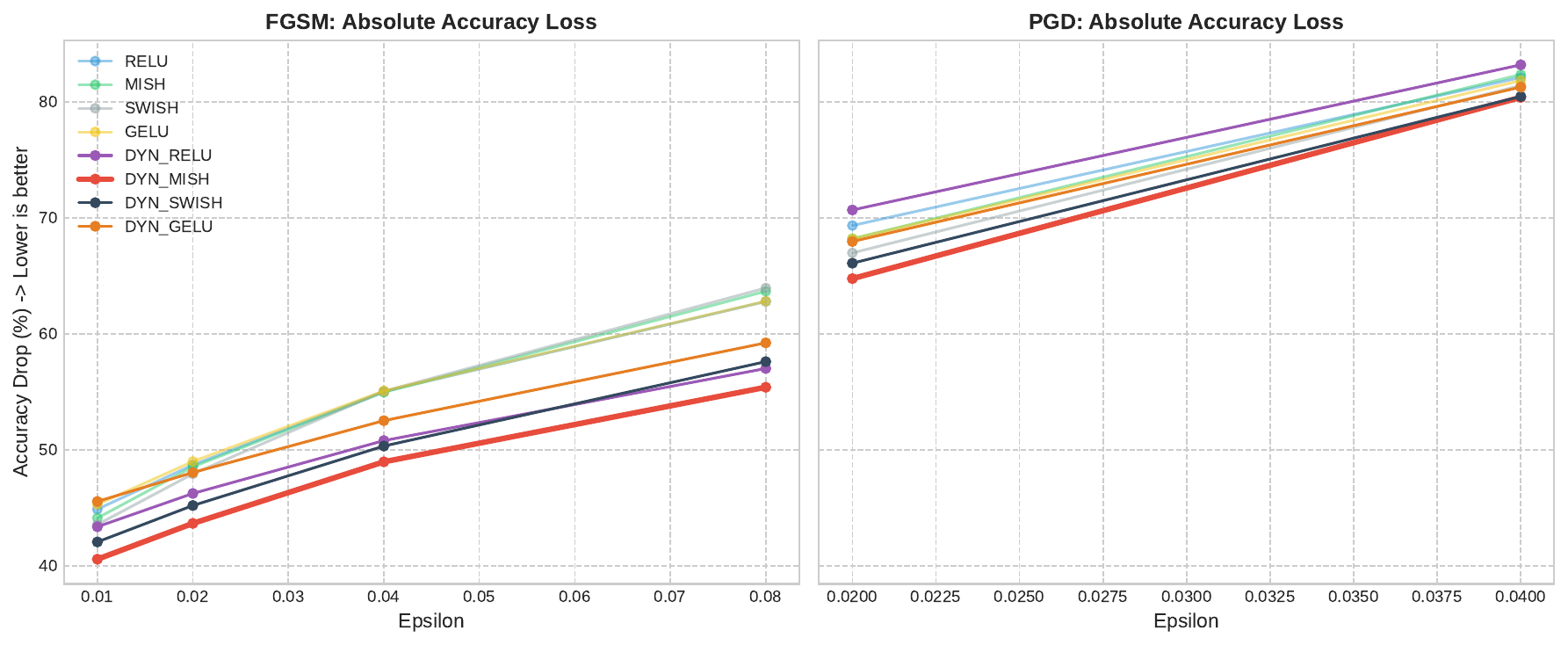}

\caption{Adversarial robustness under FGSM and PGD.}
\label{fig:unified_adversarial}
\end{figure}

Figure~\ref{fig:unified_adversarial} reports accuracy drop under FGSM~\cite{Goodfellow2015FGSM} and PGD~\cite{Madry2018PGD} attacks at multiple perturbation budgets on CIFAR-10/ResNet18.
Under FGSM at $\varepsilon = 0.01$, \dynactmish{} achieves the lowest accuracy drop at $40.56\%$, compared to $45.29\%$ for GELU~\cite{Hendrycks2016GELU}, a $4.73$ pp advantage.
At $\varepsilon = 0.08$, \dynactmish{} sustains a $55.39\%$ drop versus $62.79\%$ for ReLU ($7.40$ pp advantage), $63.67\%$ for Mish ($8.28$ pp), and $63.95\%$ for Swish ($8.56$ pp).
Under PGD at $\varepsilon = 0.02$, \dynactmish{} drops $64.76\%$ versus $69.34\%$ for ReLU, a $4.58$ pp margin.
At PGD $\varepsilon = 0.04$, \dynactmish{} achieves $80.39\%$ drop versus $82.11\%$ for ReLU ($1.72$ pp) and $82.36\%$ for Mish ($1.97$ pp).
These results indicate that \dynactmish{} is the most adversarially robust variant, with the largest margins under high-budget FGSM attacks and consistent but smaller improvements under PGD.

\subsubsection{Regularization ablation}

The no-penalty baseline achieves $79.82\%$ accuracy.
L1 regularization at $\lambda = 10^{-5}$ yields $78.91\%$, improving to $79.11\%$ at $\lambda = 10^{-4}$ and peaking at $80.02\%$ at $\lambda = 10^{-3}$.
L2 follows a flatter trajectory: $79.80\%$ at $\lambda = 10^{-5}$, $79.34\%$ at $\lambda = 10^{-4}$, and $79.92\%$ at $\lambda = 10^{-3}$.
The best configuration (L1, $\lambda = 10^{-3}$) achieves $+0.20$ pp over the baseline, with learned parameters $\bar{\alpha} = 0.1319$ and $\bar{\beta} = 0.0124$, indicating substantial parameter shrinkage toward near-linear behavior.
The accuracy optimum at moderate regularization preserves parameter diversity while providing a small but consistent accuracy improvement, suggesting that sparse regularization toward near-linear shapes is beneficial but that collapsing the parameters fully overshoots the optimum.

\begin{figure}[H]
\centering
\includegraphics[width=\linewidth]{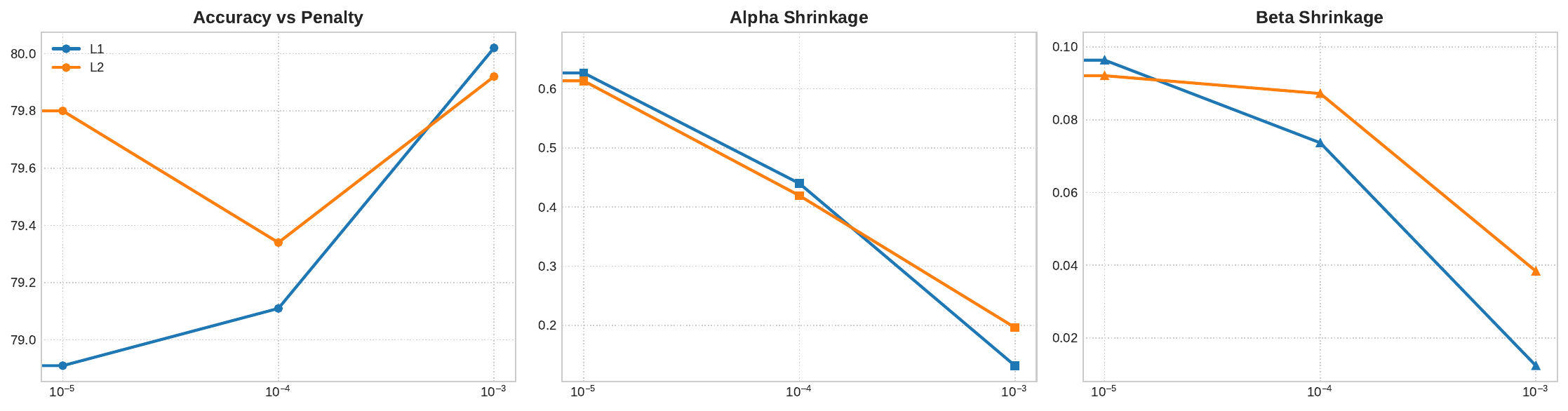}
\caption{L1 and L2 Regularization on dynActivation Parameters.}
\label{fig:reg_shrink}
\end{figure}

\subsection{Efficiency tests}

\subsubsection{Convergence test}

\begin{figure}[H]
\centering
\includegraphics[width=0.65\linewidth]{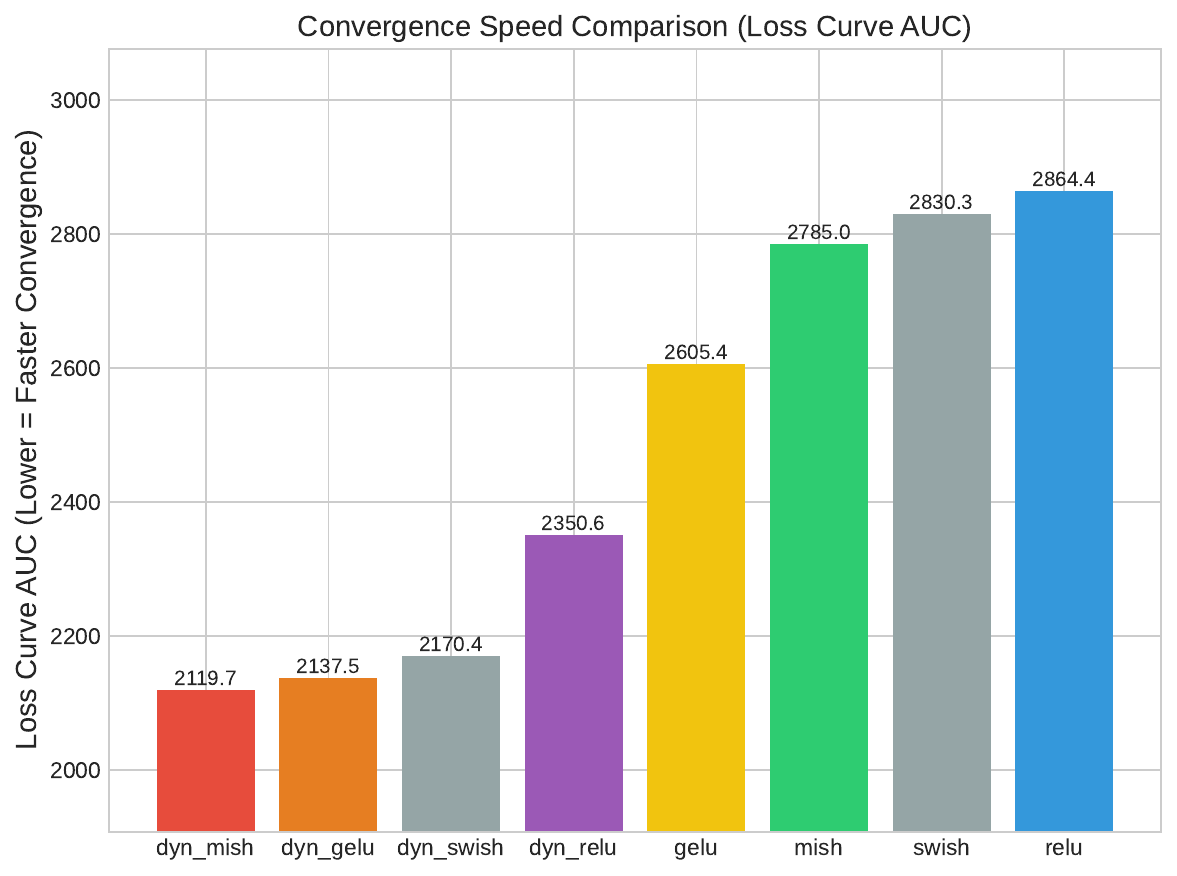}
\caption{AUC-based convergence comparison.}
\label{fig:unified_convergence}
\end{figure}

Table~\ref{tab:theory_convergence} summarizes the quantitative convergence comparison. 
\dynactmish{} achieves the lowest AUC at 2120, a $24\%$ reduction relative to static Mish (2785) and a $26\%$ reduction relative to static ReLU (2864).
\dynactgelu{} (AUC 2138) and \dynactsilu{} (AUC 2170) also substantially outperform all static baselines.
On threshold-based metrics, \dynactgelu{} reaches training loss $0.5$ fastest, requiring only 711 steps versus 781 for static GELU and ReLU.
These results confirm that dynamization consistently accelerates loss reduction, with the ordering \dynactmish{} $<$ \dynactgelu{} $<$ \dynactsilu{} $<$ \dynactrelu{} $<$ GELU $<$ Mish $<$ ReLU for the AUC metric.

\begin{table}[H]
\centering
\caption{Theory-convergence statistics across activations.}
\label{tab:theory_convergence}
\begin{tabular}{lcccc}
\toprule
Activation & AUC Loss & Steps to 1.0 & Steps to 0.5 & Conv. rate \\
\midrule
    relu & 2864.364252 & 277 & 781 & 0.000206 \\
mish & 2784.969175 & 204 & 835 & 0.000203 \\
swish & 2830.302696 & 293 & 835 & 0.000211 \\
gelu & 2605.357465 & 204 & 781 & 0.000209 \\
\dynactrelu{} & 2350.585219 & 257 & 781 & 0.000213 \\
\dynactmish{} & 2119.673199 & 226 & 781 & 0.000213 \\
\dynactsilu{} & 2170.423420 & 245 & 835 & 0.000221 \\
\dynactgelu{} & 2137.512088 & 204 & 711 & 0.000215 \\
\bottomrule
\end{tabular}
\end{table}

\subsubsection{Runtime benchmark}

Eventhough dynActivation-variants seem to provide some performance improvements, they also introduce computational overhead because of the additional parameters and gradient pathways.

\begin{figure}[H]
\vspace{-1em}

\centering
\includegraphics[width=0.86\linewidth]{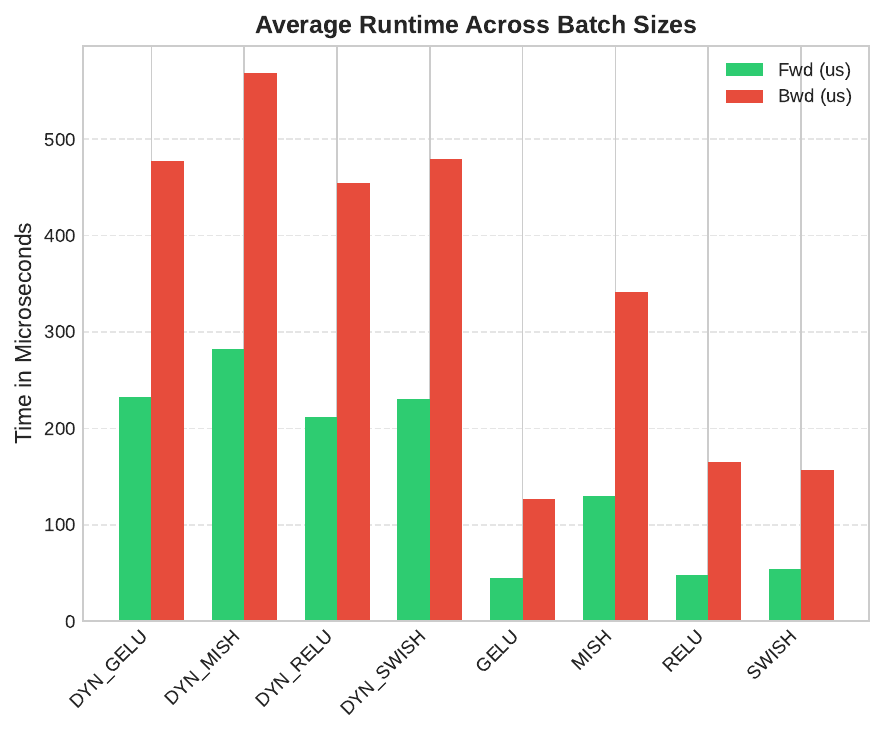}
\caption{Operator-level runtime benchmark.}
\label{fig:unified_runtime}
\vspace{-1em}

\end{figure}

Figure~\ref{fig:unified_runtime} measures forward and backward times across device types, numeric precision, batch size, and input dimension.

\subsubsection{Training-efficiency test}

\begin{figure}[H]
\vspace{-1em}
\centering
\includegraphics[width=0.86\linewidth]{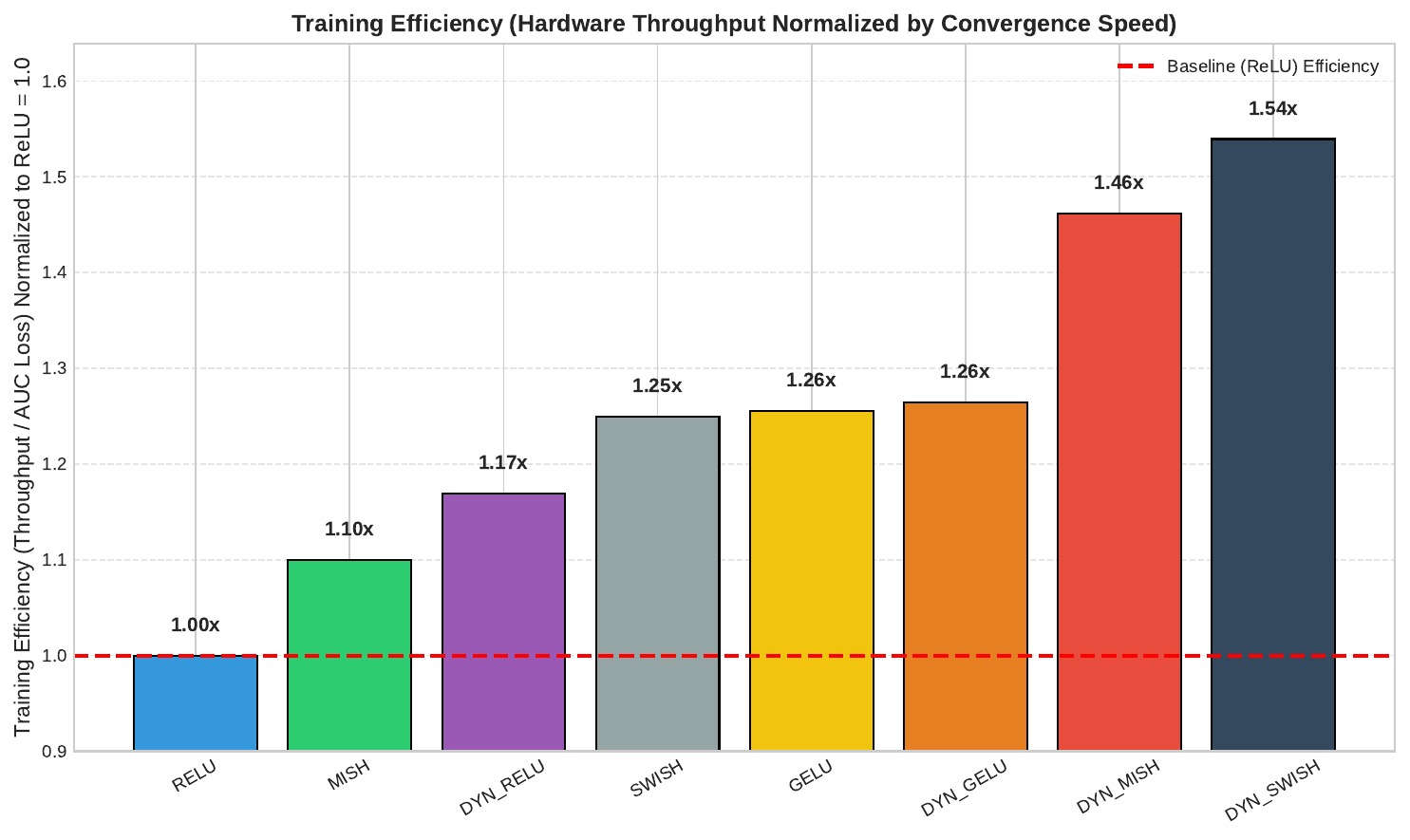}
\caption{Training efficiency combining convergence and compute cost.}
\label{fig:unified_training_eff}
\vspace{-1em}

\end{figure}

But the runtime has to be seen in relation with the convergence rate of the model to allow a conclusion if the computational overhead is worth it. Figure~\ref{fig:unified_training_eff} combines convergence behavior with hardware-level throughput. This graphic shows that the increased convergence speed seems to outweigh the computational overhead and suggests that the dynActivation variants are up to $54\%$ more efficient to train than normal ReLU.

\section{Discussion}

\textbf{CIFAR classification.}
The $+14.02$ pp improvement on CIFAR-10/AttentionCNN (from $64.25\%$ Mish to $78.27\%$ \dynactmish{}) and the $+12.39$ pp gain on CIFAR-100/AttentionCNN (from $33.31\%$ ReLU to $45.70\%$ \dynactgelu{}) are the largest individual gains observed.
These margins suggest that some architectures are particularly constrained by a fixed activation; adding two learnable parameters per layer is sufficient to substantially improve performance.
Across the broader 27-activation sweep, dynActivation(Mish) places first at $79.72\%$. The unified paired testing shows that this advantage is statistically significant over its static counterpart Mish ($p = 0.0375$), as well as over established standards like ReLU ($p = 0.0205$) and SELU ($p = 4.20 \times 10^{-5}$). The only activations performing statistically on par with the leading variant are other members of the dynActivation family (GELU, SiLU).

\textbf{Depth scaling.}
The fact that \dynact{} is the only activation maintaining $>95\%$ accuracy from 1 to 75 layers---while all static baselines collapse below $60\%$ before layer 50---cannot be attributed to a simple accuracy offset.
The layer-wise shape analysis explains the mechanism: the network learns to reduce $\alpha$ and increase $\beta$ in deep layers, producing near-identity activations ($\alpha \approx \beta \approx 0.5$) that act as effective skip connections without architectural modification.
This transition from $\alpha \approx 1.12$ in early layers to $\alpha \approx 0.45$--$0.50$ in layers 30--45 is a continuous, task-driven interpolation between nonlinear and linear behavior.

\textbf{Learned parameter trajectories and selective nonlinearity.}
The parameter trajectory (Figure~\ref{fig:unified_monotonicity}) confirms that $\bar{\alpha}$ converges from $0.90$ to $0.75$ by epoch 10, while $\bar{\beta}$ stabilizes at $-0.08$ by epoch 5.
The converged values correspond to a slightly attenuated Mish nonlinearity with a small negative linear correction, a regime that the network consistently selects across architectures.
The large inter-layer variance in $\beta$ ($\pm 0.2$) versus the smaller variance in $\alpha$ ($\pm 0.10$) indicates that the linear-path parameter is the primary instrument of layer-wise specialization, while the nonlinearity-scaling parameter converges to a more uniform global value.

\textbf{Adversarial robustness and convergence.}
The adversarial advantage under FGSM~\cite{Goodfellow2015FGSM} at $\varepsilon = 0.08$---where \dynactmish{} reduces accuracy drop by $7.40$ pp relative to ReLU~\cite{nair2010relu} ($55.39\%$ vs.\ $62.79\%$), $8.28$ pp relative to Mish, and $8.56$ pp relative to Swish---is consistent with the hypothesis that a smoother, partially linearized activation reduces the network's local sensitivity to input perturbations.
The $24\%$ convergence-AUC reduction (2120 vs.\ 2785 for Mish) indicates that the gradient advantage observed in 50-layer networks translates into a practical training-time benefit under standard experimental protocols.

\textbf{LLM transfer.}
The $10.3\%$ relative perplexity reduction after 5{,}620 steps provides evidence that the dynActivation idea is not image-specific.
The near-zero residual gap at 34{,}300 steps ($3.806$ vs.\ $3.809$) is consistent with asymptotic equivalence rather than a long-run regression, and the short-run advantage may be practically important in low-compute training regimes.

\textbf{Optimizer and initialization stability.}
Across 27 optimizer--init--learning-rate configurations, \dynactmish{} achieves the highest mean accuracy ($52.81\%$), leading Mish by $+3.34$ pp and ReLU by $+10.56$ pp. The best single configuration (kaiming\_normal, RMSProp, $\text{lr}=0.001$) reaches $75.77\%$. This consistent advantage suggests that dynActivation is a robust drop-in replacement rather than a method requiring careful co-optimization of the training protocol.

\textbf{Distribution-shift robustness.}
The distribution-shift evaluation~\cite{Hendrycks2019Robustness} reveals a mixed rather than neutral picture: \dynactmish{} leads under Gaussian noise (${+}2.42$ pp over ReLU at severity 1) and brightness corruptions (${+}4.02$ pp over ReLU at severity 5), but falls behind under contrast perturbations ($-4.38$ pp vs.\ Swish at severity 1). This corruption-type-dependent profile warrants further investigation into whether the learned parameter regime can be regularized toward more uniform robustness.

\section{Limitations}

Several limitations qualify the claims made in this paper.
First, while dynActivation(Mish) achieves the highest mean accuracy overall, its advantage over Apa is not yet statistical significant in the aggregated paired setting. Further testing is needed to proof the statistical significance of the empirical results.

Second, the LLM perplexity advantage of $10.3\%$ after 5{,}620 steps collapses to $0.09\%$ at 34{,}300 steps, making the long-run benefit effectively zero; dynActivation in language models should therefore be framed as a convergence accelerator rather than an asymptotic improvement.

Third, while the two additional parameters per layer are lightweight, the computational overhead relative to static activations has not been eliminated; the runtime benchmark quantifies this cost and it must be weighed against the accuracy gains in latency-sensitive applications.

Fourth, the current understanding of why $\bar{\alpha}$ and $\bar{\beta}$ converge to these specific values rests on an empirical observation rather than a theoretical guarantee; a principled analysis of the optimization landscape that explains these attractors is left for future work.

\appendix

\section*{Use of generative AI}
Generative AI tools, primarily Perplexity (GPT-5.4 and Gemini 3.1 Pro), were used as support tools for language refinement, drafting assistance, preparation of selected figures and tests. Any AI-generated suggestions or draft materials were critically reviewed, revised, and verified by the author. All scientific reasoning, experimental design, analysis, and final interpretation were conducted and verified by the author. The author takes full responsibility for the content of this manuscript.

\end{document}